\documentclass{article} 
\usepackage{iclr2027_conference,times}

\usepackage{amsmath,amsfonts,bm}

\def\eqref#1{equation~\ref{#1}}

\def\1{\bm{1}}

\DeclareMathAlphabet{\mathsfit}{\encodingdefault}{\sfdefault}{m}{sl}
\SetMathAlphabet{\mathsfit}{bold}{\encodingdefault}{\sfdefault}{bx}{n}

\usepackage{hyperref}
\usepackage{url}
\usepackage{graphicx}

\usepackage{tikz}
\usepackage{booktabs}
\usepackage{tabularx}
\usepackage{array}
\usepackage{multirow}
\usepackage{colortbl}
\usepackage{amssymb} 
\usepackage{graphicx}
\usepackage{calc}
\usepackage{capt-of}
\usepackage{makecell}
\usepackage{wrapfig}
\usepackage[table]{xcolor}
\providecolor{tableblue}{RGB}{235,244,252}

\newsavebox{\ablationbox}
\newsavebox{\interventionbox}

\newcolumntype{Y}[1]{%
    >{\hsize=#1\hsize\centering\arraybackslash}X%
}

\newcolumntype{C}[1]{>{\centering\arraybackslash}m{#1}}

\newcommand{\taskanchorpi}{\textsc{TaskAnchor}-$\pi_{0.5}$}
\newcommand{\taskanchorpibf}{%
    \textbf{\textsc{TaskAnchor}}-$\boldsymbol{\pi_{0.5}} \ $%
}

\DeclareRobustCommand{\redcircle}{%
  \tikz[baseline=-0.55ex]
  \draw[red!80!black, dashed, line width=0.7pt] (0,0) circle (0.75ex);%
}
\DeclareRobustCommand{\greencircle}{%
  \tikz[baseline=-0.55ex]
  \draw[green!50!black, dashed, line width=0.7pt] (0,0) circle (0.75ex);%
}

\title{\textsc{TaskAnchor}: Grounding Task State in Reactive VLAs for Long-Horizon Manipulation}

\author{
\textbf{Hengyan Liu}$^{1,4*}$ \quad
\textbf{Wenlve Zhou}$^{2*}$ \quad
\textbf{Bo Yue}$^{1,4*}$ \quad
\textbf{Yongyi Su}$^{2}$ \quad
\textbf{Ruixiang Wang}$^{1,4}$\\
\textbf{Zhanqi Zhang}$^{4}$ \quad
\textbf{Dekun Lu}$^{4}$ \quad
\textbf{Wei Gao}$^{4}$ \quad
\textbf{Xiaofen Xing}$^{3\dagger}$ \quad
\textbf{Kui Jia}$^{1,4\dagger}$\\[3pt]
$^{1}$The Chinese University of Hong Kong, Shenzhen
\quad
$^{2}$Foshan University\\
$^{3}$South China University of Technology
\quad
$^{4}$DexForce\\[3pt]
\texttt{xiaofenxing@scut.edu.cn}
\quad
\texttt{kuijia@cuhk.edu.cn}\\[3pt]
$^{*}$Equal contribution. Wenlve Zhou led the project.
\quad
$^{\dagger}$Corresponding authors.
}

\makeatletter
\@ifpackageloaded{wrapfig}{}{\usepackage{wrapfig}}
\@ifpackageloaded{placeins}{}{\usepackage{placeins}}
\@ifpackageloaded{needspace}{}{\usepackage{needspace}}
\makeatother

\iclrfinalcopy 
\begin{document}

\maketitle

\begin{abstract}
Reactive vision-language-action (VLA) policies suffer from \emph{task-state aliasing} in long-horizon manipulation, where identical multimodal inputs call for distinct, context-dependent actions. Given that pretrained VLAs already possess rich control primitives to express diverse behaviors, we hypothesize that the execution bottleneck lies not in policy capacity, but in input ambiguity. In this paper, we propose \textsc{TaskAnchor}, a lightweight adapter that grounds task state by injecting execution context into the VLA’s native input space.
During post-training, \textsc{TaskAnchor} learns to represent the semantic execution stage as a milestone-supervised coordinate prepended to the language instruction, while incorporating fine-grained historical evidence via a residual update to the current visual tokens. This formulation avoids generating complex subtask instructions and leaves the backbone architecture unchanged.
Across long-horizon benchmarks, \textsc{TaskAnchor} delivers substantial gains, achieving approximately $6\times$ the average success rate of the $\pi_{0.5}$ and X-VLA baselines on RMBench and more than doubling the task success rate of $\pi_{0.5}$ on RoboMemArena.
Real-robot experiments further validate reliable multi-stage execution, with the same policy adapting its subsequent behaviors using earlier human interactions as in-context cues. Our project website is available at
{https://taskanchor.netlify.app/}.
\end{abstract}

\vspace{-0.22in}
\section{Introduction}
\label{sec:introduction}
\vspace{-0.12in}

Vision--language--action models (VLAs)~\citep{zheng2026x, pi05, wu2026pragmatic, team2026xiaomi, wang2026qwen, zhou2026dexterity} build on vision--language pretraining~\citep{beyer2024paligemma, marafioti2025smolvlm, dang2026rynnbrain} to learn generalizable manipulation policies from large-scale embodied trajectory data~\citep{khazatsky2024droid, wu2024robomind, grauman2022ego4d}. Most existing VLAs adopt a reactive policy formulation, predicting short-horizon action chunks conditioned on the current observation and task instruction. In long-horizon manipulation~\citep{chen2026rmbench, lei2026robomemarena}, however, appropriate action selection depends on historical context that instantaneous inputs cannot capture. This reliance creates severe \emph{task-state aliasing}: visually similar states under identical instructions dictate conflicting behaviors, causing reactive policies to routinely fail even when equipped with the required manipulation skills.

\begin{figure*}[htbp]
\centering
\vspace{-0.12in}
\includegraphics[width=\textwidth]{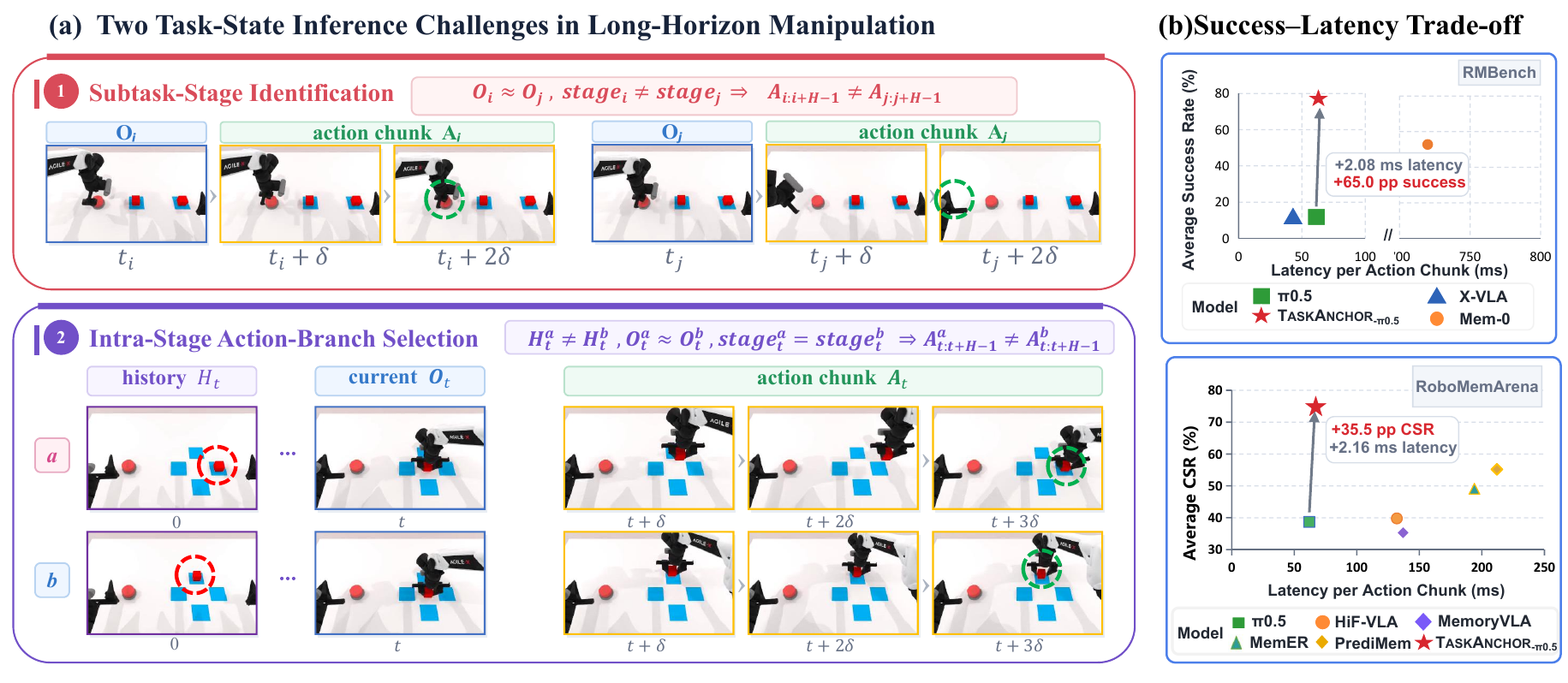}
\vspace{-0.35in}
\caption{\textbf{Task-state aliasing and \textsc{TaskAnchor} performance.}
\textbf{(a)} In long-horizon manipulation, action selection depends on the execution stage (top) and on earlier interactions within the same stage (bottom).
Dashed circles mark historical cues (\redcircle) and their downstream effects (\greencircle).
\textbf{(b)} Compared with existing methods, \textsc{TaskAnchor} improves long-horizon performance on both benchmarks with minimal additional latency overhead.}
\label{fig:task_state_tradeoff}
\vspace{-0.2in}
\end{figure*}
We describe this decision-critical context as the \emph{task state}. Inferring the task state requires resolving two complementary aspects of history: identifying the current subtask stage~\citep{liu2026palm, li2025scalable}, which determines where the robot is in the overall task, and recovering history-grounded task context~\citep{shi2026memoryvla, hu2026resolving}, which determines what the current observation implies and which action should be executed (Figure~\ref{fig:task_state_tradeoff}(a)).
Consider a robot instructed to move a red block onto a blue pad and then return it to its original slot. There are two identical slots, only one of which initially contains the block. Just before placing the block on the pad and just after picking it up again, the robot holds the block in the same pose above the pad. In both cases, the pad and both slots are empty, yielding virtually identical observations. Yet the next action differs: the robot should first lower the block onto the pad, but after retrieval, carry it back toward the slot. Recognizing the return stage, however, does not determine which slot to use. With both slots now empty, selecting the correct destination requires remembering whether the block was initially picked up from the above or right slot~\citep{chung2026rethinking}.


Existing approaches address parts of this spectrum, but struggle to resolve both forms simultaneously. History-aware policies~\citep{shi2026memoryvla, koo2026hamlet, shi2026memoryvla++, li2026remem} retain temporal evidence to aid intra-stage action branching, yet leave the subtask stage implicit, forcing the policy to unreliably infer stage transitions from raw latent states. Hierarchical approaches~\citep{chen2026rmbench, lei2026robomemarena, sridhar2026scaling} make subtask selection explicit through a separate VLM planner~\citep{bai2025qwen3, zhang2026harness} that generates instructions for a low-level policy ~\citep{pi05, zheng2026x}. 
Such fractured architectures discard fine-grained visual details and incur high inference latency.
These approaches motivate a task-state interface that predicts the current subtask stage in a more compact way while preserving the historical context needed for selecting intra-stage actions.

Reactive VLAs naturally lend themselves to this interface: their prompt responsiveness shows that native inputs effectively steer learned manipulation skills~\citep{brohan2023rt, kim2024openvla}. We therefore hypothesize that \textit{grounding these inputs in decision-relevant history elicits latent long-horizon capabilities}. Guided by this hypothesis and the task-state decomposition above, we introduce \textsc{TaskAnchor}, a lightweight adaptation framework that equips pretrained reactive VLAs~\citep{pi05, zheng2026x} with explicit task-state awareness. \textsc{TaskAnchor} resolves history-dependent action ambiguity through a history-conditioned visual residual and grounds action generation in the current execution stage through a milestone-supervised task-state coordinate.
These complementary signals are injected through the native multimodal interfaces of the VLA, without introducing a separate planner, subtask instructions, or modifying the action-generation mechanism.

We evaluate \textsc{TaskAnchor} across complex long-horizon tasks of RMBench~\citep{chen2026rmbench} and RoboMemArena~\citep{lei2026robomemarena}. As detailed in Figure~\ref{fig:task_state_tradeoff}(b), 
\textsc{TaskAnchor}-$\pi_{0.5}$ yields consistent performance gains, improving the average success rate on RMBench by an absolute $+0.650$ and the average cumulative success rate (CSR) on RoboMemArena by $35.5$ percentage points (from $38.7\%$ to $74.2\%$).
Remarkably, this adaptation incurs only approximately $2\,\mathrm{ms}$ of additional inference latency per action chunk. Further probing indicates that the inferred task-state representations systematically modulate the action selection of the pretrained policy, substantiating our hypothesis that pretrained VLAs possess latent long-horizon capabilities which can be unlocked through lightweight disambiguation.

Our contributions are summarized as follows:
\begin{itemize}
\item We reveal that long-horizon failures in reactive VLAs stem from conflating two distinct ambiguities. By decoupling task state into a task-state coordinate and granular historical details, we transform task-state inference from an intractable temporal modeling problem into two targeted conditioning tasks.

\item We resolve task-state aliasing entirely within the VLA's native input space. By injecting task-state coordinates as augmented language instructions and intra-stage action-selection-relevant historical evidence as visual residuals, \textsc{TaskAnchor} requires zero changes to the underlying VLA architecture or action-generation mechanism.

\item We demonstrate that pretrained VLAs possess long-horizon capabilities once properly grounded with task state. With only 27M additional parameters and approximately $2\,\mathrm{ms}$ latency overhead, \textsc{TaskAnchor} achieves a $\sim$$6\times$ increase in average success rate on RMBench, more than $2\times$ the performance of $\pi_{0.5}$ on RoboMemArena, and reliable multi-stage real-robot execution through history-aware in-context adaptation. 
\end{itemize}

\vspace{-0.2in}
\section{Related Work}
\vspace{-0.1in}

\vspace{-0.02in}\textbf{Pretrained Vision--Language--Action Models.}
Vision--language--action models combine vision--language pretraining with robot trajectories to unify semantic grounding and visuomotor control. RT-2 represents actions as language-model tokens to transfer Internet-scale visual--semantic knowledge to robotics~\citep{brohan2023rt}, while Octo, OpenVLA, and X-VLA learn generalist policies from large-scale cross-embodiment data~\citep{team2024octo,kim2024openvla, zheng2026x}. The $\pi$ series instead couples a pretrained vision--language backbone with a flow-matching action expert for continuous control~\citep{pi0,pi05}. Despite their rich semantic, spatial, and behavioral priors, these models are typically deployed reactively without explicit task-state conditioning, making visually similar stages prone to repeated actions or premature termination. \textsc{TaskAnchor} addresses this limitation through lightweight post-training that conditions action generation on task state.

\vspace{-0.02in}\textbf{Temporal Context for Long-Horizon Control.}
Previous works incorporate temporal context through frame stacking, attention, recurrent states, or compact memory. RoboFlamingo uses sequential visual context~\citep{li2023roboflamingo}; HAMLET introduces moment tokens and lightweight memory~\citep{koo2026hamlet}; MemoryVLA selectively retrieves task-relevant observations from a perceptual-cognitive memory bank~\citep{shi2026memoryvla}; and Long-VLA exploits task phases through phase-aware input masking~\citep{fan2025long}; RoboTTT compresses long visuomotor histories into recurrent fast weights through test-time training layer~\citep{jiang2026robottt}. These methods establish the importance of history for non-Markovian dependencies and long-horizon skill composition. Rather than directly exposing long histories to the action generator, \textsc{TaskAnchor} distills task-relevant evidence into a residual correction of the current visual representation and a compact task-state coordinate, targeting stage-level observation aliasing rather than merely expanding the temporal receptive field.

\vspace{-0.02in}\textbf{High-Level Task Planning and Explicit Reasoning.}
Another line of work represents long-horizon structure through planning or semantic reasoning. SayCan and Inner Monologue generate skill sequences or intermediate decisions using language models and low-level policies~\citep{ahn2022can,huang2022inner}. More recent methods, including $\pi_{0.5}$, LoHoVLA, VLA-OS, Mem-0, PrediMem, and CoT-VLA, explore semantic subtasks, hierarchical execution, visual plans, or intermediate reasoning traces~\citep{pi05,yang2025lohovla,gao2026vla,chen2026rmbench,lei2026robomemarena, zhao2025cot}. These approaches organize behavior around explicit subtask or goal representations. In contrast, \textsc{TaskAnchor} generates neither free-form subtasks, future trajectories, nor candidate plans, and introduces no separate planning loop. Instead, it investigates whether a low-bandwidth task-state coordinate can select stage-consistent behavior from a pretrained VLA, thereby activating existing behavioral knowledge without specifying the next action semantically.



\vspace{-0.13in}
\section{Method}
\label{sec:method}
\vspace{-0.12in}

\begin{figure*}[!t]
\centering
\vspace{-0.1in}
\includegraphics[width=5.5in]{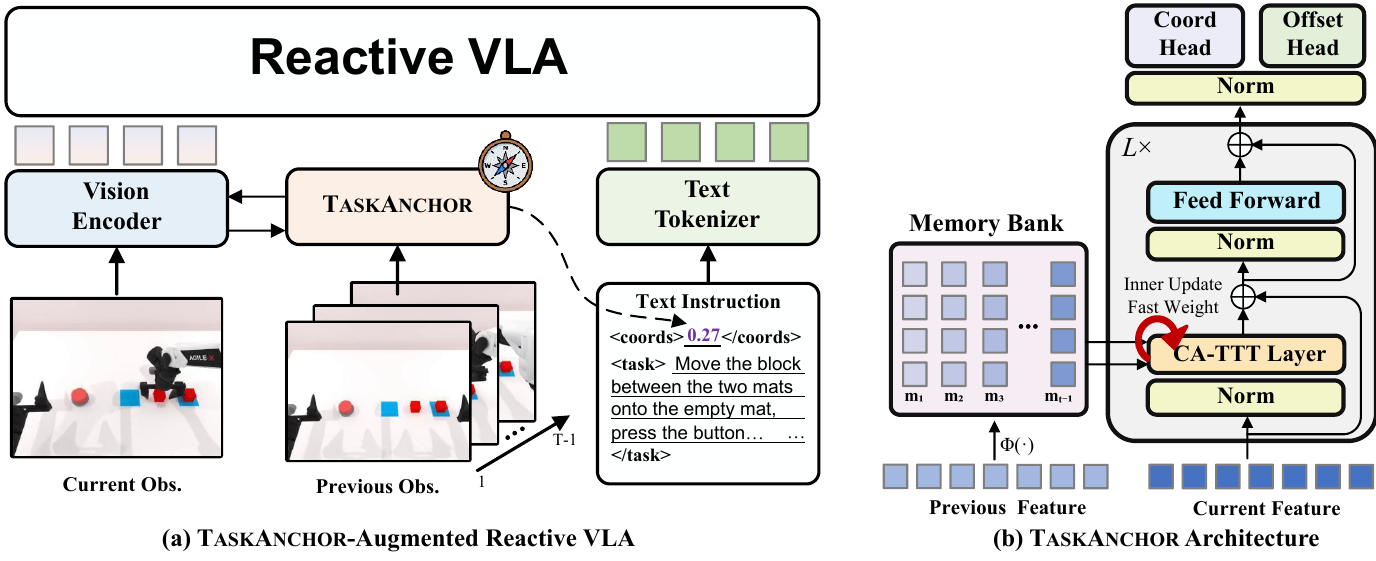}
\vspace{-0.27in}
\caption{
\textbf{Overview of \textsc{TaskAnchor}.}
\textbf{(a)} \textsc{TaskAnchor} augments a pretrained reactive VLA with memory-grounded task-state conditioning. Using the current observation and execution history, it predicts a coordinate and a residual visual offset. These outputs condition the reactive VLA through its native language and visual interfaces, respectively.
\textbf{(b)} Given current visual features and a compact memory constructed from historical features using $\Phi(\cdot)$, \textsc{TaskAnchor} uses $L$ layers to model temporal interactions. Two output heads then predict the coordinate and visual offset.
}\vspace{-0.16in}
\label{fig:pipeline}
\end{figure*}

We present \textsc{TaskAnchor}, a lightweight framework that enhances reactive VLAs for long-horizon manipulation while preserving their original action-generation paradigm. As shown in Figure~\ref{fig:pipeline}, the system leverages a compact memory buffer to extract temporal evidence for dual-pathway modulation, yielding refined visual tokens for intra-stage dynamics and an evolving task-state coordinate for inter-stage progression. Injecting these signals directly into native visual and language modalities of the VLA steers the action distribution toward long-horizon goals while leaving the generation architecture intact.

\vspace{-0.12in}
\subsection{Task-State Policy Formulation}
\label{sec:task_state_formulation}
\vspace{-0.09in}

Let $X_t = E_{\mathrm{vis}}(O_t)$ denote the tokens produced by the visual encoder $E_{\mathrm{vis}}(\cdot)$ from observation $O_t$.
A reactive VLA conditions action selection only on instantaneous inputs, predicting an $H$-step action chunk $A_{t:t+H-1}$~\citep{zhao2023learning} according to $p_\theta(\cdot \mid X_t,\ell)$ given instruction $\ell$. Relying exclusively on this instantaneous mapping induces task-state aliasing in long-horizon tasks.

Given a compact visual memory $\mathcal{M}_t$ of observations before time $t$, \textsc{TaskAnchor} infers an augmented task state $C_t=(\widetilde{X}_t,p_t)$. Here, $\widetilde{X}_t$ denotes history-conditioned visual tokens obtained by refining the current visual representation $X_t$, while $p_t\in[0,1]$ is the task-state coordinate, a scalar representing the semantic stage of task execution rather than a spatial position. The coordinate is further incorporated into the original instruction $\ell$ through a fixed prompt template (Figure~\ref{fig:pipeline}(a)), yielding the augmented instruction $\ell_t^{+}$. The reactive VLA then predicts actions from the task-state-conditioned inputs $(\widetilde{X}_t,\ell_t^{+})$.

Here, $\theta^\prime$ denotes the VLA parameters after joint post-training with the \textsc{TaskAnchor} on the target tasks. Separating task-state inference from action generation gives the joint factorization
\begin{align}
    p_{\theta^\prime,\psi}
    \left(
        A_{t:t+H-1},C_t
        \mid
        X_t,\mathcal{M}_t,\ell
    \right)
    \;=\;
    \underbrace{
        p_{\theta^\prime}
        \left(
            A_{t:t+H-1}
            \mid
            \widetilde{X}_t,\ell_t^{+}
        \right)
    }_{\text{Reactive VLA}}
    \cdot
    \underbrace{
        p_{\psi}
        \left(
            C_t
            \mid
            X_t,\mathcal{M}_t
        \right)
    }_{\textsc{TaskAnchor}},
    \label{eq:joint_factorization}\vspace{-0.2in}
\end{align}
The \textsc{TaskAnchor} handles history-based task-state inference, while the VLA generates actions through its native visual and language interfaces.

\vspace{-0.12in}
\subsection{Context-Adaptive Visual Conditioning}
\vspace{-0.08in}
\label{sec:context_adaptive_visual}

\paragraph{TTT as Implicit Memory.}
Test-Time-Training (TTT)~\citep{sun2024learning, han2026vit} layer  compresses sequence history into the weights of a lightweight inner model $g(\cdot;\omega)$ through self-supervised updates. Instead of explicitly storing historical tokens as a KV cache~\citep{vaswani2017attention}, TTT encodes the context into fast weights, allowing current queries to retrieve historical information through a forward pass. This implicit memory representation avoids the quadratic growth of self-attention with sequence length while maintaining a fixed-size parametric state.

\vspace{-0.1in}\paragraph{Context-adaptive TTT (CA-TTT).}
We reinterpret TTT as a transient history-conditioned adapter rather than a recurrent parameter state. Based on this view, we develop CA-TTT with $L$ stacked layers to extract history-conditioned visual features. Unlike online finetuning, CA-TTT does not update persistent model parameters during inference. Unlike recurrent TTT~\citep{sun2024learning}, where adapted weights are propagated across sequential inputs, CA-TTT reconstructs temporary fast weights independently at each decision step $t$ from an offline-learned initialization $\omega^{(m)}$. These fast weights are used only for current-step feature extraction and discarded after readout.

Given the current observation $O_t$, we maintain a chronological memory bank of compressed historical features:
$\mathcal{M}_t = [\mathbf{m}_1; \ldots; \mathbf{m}_{t-1}]$, where $\mathbf{m}_i = \Phi(X_i)$. Spatial compression $\Phi(\cdot)$ uses bilinear interpolation to downsample the visual token grid, reducing the number of stored tokens, while rotary position embeddings encode temporal order within the CA-TTT layers. At layer $m \in \{1,\ldots,L\}$, layer-specific projections map the shared memory bank to keys and values:
\begin{align}
    K_t^{(m)}
    &= f_k^{(m)}(\mathcal{M}_t),
    \qquad
    V_t^{(m)} = f_v^{(m)}(\mathcal{M}_t).
    \label{eq:layerwise_history_kv}
\end{align}
For nonempty memory, a lightweight MLP $g(\cdot;\omega)$ learns the historical key--value association through a self-supervised~\citep{liu2021self} reconstruction objective:
\begin{align}
    \mathcal{L}_{\mathrm{TTT},t}^{(m)}(\omega)
    &= \frac{1}{N_t}
    \sum_{j=1}^{N_t}
    \left\|
        g\left(K_{t,j}^{(m)};\omega\right)
        -
        V_{t,j}^{(m)}
    \right\|_2^2,
    \label{eq:ttt_objective}
\end{align}
where $N_t$ denotes the number of historical memory tokens.Starting from the offline-learned initialization $\omega^{(m)}$, CA-TTT performs a single fast-weight update:
\begin{align}
    \widehat{\omega}_t^{(m)}
    &= \omega^{(m)}
    - \eta_{\mathrm{ttt}}
    \left.
        \nabla_{\omega}
        \mathcal{L}_{\mathrm{TTT},t}^{(m)}(\omega)
    \right|_{\omega=\omega^{(m)}}.
    \label{eq:layerwise_fast_weight_update}
\end{align}
Here, $\eta_{\mathrm{ttt}}$ is the inner-loop learning rate. For simplicity, the update is written in its standard gradient-descent form; the implementation applies gradient normalization for stable fast-weight adaptation. The resulting computation depends on the memory size $N_t$, but does not introduce a persistent optimization state across decisions.


\vspace{-0.05in}\paragraph{Query readout.}
The query stream is initialized as $Q_t^{(0)} = [q_p;\, f_q(X_t)]$, where $q_p$ is a learnable task-state coordinate query and $f_q$ projects current visual features $X_t$. The query length remains independent of the number of historical observations. At layer $m$, current queries retrieve history-conditioned features from the temporary fast weights through a residual update:
\begin{equation}
    \overline{Q}_t^{(m)}
    =
    Q_t^{(m-1)}
    +
    g\left(
        Q_t^{(m-1)};
        \widehat{\omega}_t^{(m)}
    \right).
    \label{eq:ca_ttt_query_readout}
\end{equation}
This operation serves as a form of \emph{implicit cross-attention}: $Q_t^{(m-1)}$, $K_t^{(m)}$, and $V_t^{(m)}$ play roles analogous to queries, keys, and values, while retrieval is parameterized by fast weights rather than an explicit attention matrix. A residual feed-forward block then produces $Q_t^{(m)}$; layer normalization~\citep{ba2016layer} is omitted from the equations for clarity. After readout, fast weights are discarded and are not propagated to subsequent decisions. The memory bank remains the only history-dependent state, while all offline-learned parameters remain fixed during inference.

After $L$ layers, the visual query tokens $Q_{O,t} \subset Q_t^{(L)}$ are passed to an offset head $H_{\psi_o}(\cdot)$ to refine the current visual features:
\begin{equation}
    \widetilde{X}_t = X_t + H_{\psi_o}(Q_{O,t}),
\end{equation}
which provides historical conditioning through the native visual interface of the VLA.

\vspace{-0.1in}
\subsection{Task-State Coordinate Prediction and Conditioning}
\vspace{-0.05in}
\label{sec:task_coordinate}

To represent the semantic execution stage without relying on elapsed physical time, we introduce a continuous task-state coordinate $p_t^\star = m_t/M$, where $m_t\in\{0,\ldots,M-1\}$ is the index of the current semantic milestone. These milestone boundaries are readily defined by benchmark subtask annotations~\citep{chen2026rmbench, lei2026robomemarena} or extracted via temporal segmentation. At each step $t$, a lightweight coordinate head $C_{\psi_c}(\cdot)$ processes the task-state coordinate query $q_{p,t}$ extracted from the final block to predict a continuous coordinate $p_t = C_{\psi_c}(q_{p,t})$, supervised by a mean squared error loss:
\begin{equation}
    \mathcal{L}_{\mathrm{coord}} = \left\| p_t - p_t^\star \right\|_2^2.
    \label{eq:coordinate_supervision}
\end{equation}
This predicted coordinate is formatted as a normalized decimal and prepended to the language instruction $\ell$ using a fixed prompt template (Figure~\ref{fig:pipeline}(a)), yielding an augmented instruction $\ell_t^{+}$. To stabilize training via teacher forcing, we construct $\ell_t^{+}$ using the ground-truth task-state coordinate $p_t^\star$ during training, while seamlessly swapping it with the predicted $p_t$ during inference.

\vspace{-0.1in}
\subsection{Training and Inference Protocols}
\vspace{-0.05in}
\label{sec:training_inference}

The complete system—comprising the \textsc{TaskAnchor} adapter and the reactive VLA—is jointly optimized end-to-end via a composite outer objective:
\begin{equation}
    \mathcal{L} = \mathcal{L}_{\mathrm{act}} + \lambda_{\mathrm{coord}} \mathcal{L}_{\mathrm{coord}},
    \label{eq:training_objective}
\end{equation}
where $\mathcal{L}_{\mathrm{act}}$ is the VLA's native action loss and $\lambda_{\mathrm{coord}}$ balances coordinate supervision. Conditioned on $p_t^\star$ during training, the action loss $\mathcal{L}_{\mathrm{act}}$ updates the language-conditioned policy without propagating noisy gradients back through the coordinate prediction head. Note that the inner TTT loss $\mathcal{L}_{\mathrm{TTT},t}^{(m)}$ serves solely to construct step-wise fast weights and is omitted from this outer objective.

During online inference, all slow weights remain frozen. At each step $t$, the CA-TTT layers construct fast weights $\widehat{\omega}_t^{(m)}$ from history $\mathcal{M}_t$ in a single forward pass, produce the conditioning tuple $(\widetilde{X}_t, p_t)$, and discard the fast weights immediately after action execution. The compressed visual feature $\mathbf{m}_t = \Phi(X_t)$ is then appended to $\mathcal{M}_t$ for subsequent steps, with the memory bank being reset at episode boundaries.

\vspace{-0.1in}
\section{Experiments}
\vspace{-0.1in}\label{sec:experiments}


\textsc{TaskAnchor} is evaluated on simulated long-horizon benchmarks and real-world robotic tasks. It is implemented as a lightweight adapter to pretrained VLAs while retaining their action-generation mechanisms. The primary implementation builds on pretrained $\pi_{0.5}$ (\taskanchorpi{}); the same adapter design is applied to X-VLA (\textsc{TaskAnchor}-XVLA) to assess generalization across architectures. History-conditioned visual adaptation provides evidence from execution history, while predicted task-state coordinate indicates the current semantic execution stage. Neither signal requires ground-truth annotations at inference. The experiments examine the learned task-state interface, its effect on the VLA backbone, and practical robustness through the following questions:

\noindent\textbf{Q1 (Disambiguation Capability):} Can a lightweight, explicit task-state representation effectively elicit behavioral priors in pretrained VLAs to resolve task-state aliasing in long-horizon manipulation? (\emph{Sec.~\ref{sec:rmbench} \& Sec.~\ref{sec:robomemarena}})
    
\noindent\textbf{Q2 (Mechanism \& Synergy):} How do the dual task-state pathways, i.e., history-conditioned visual adaptation via fast-weights and task-state coordinate prompting, synergize to guide policy execution, and does the predicted coordinate serve as a controllable causal anchor? (\emph{Sec.~\ref{sec:task_state_analysis}})
    
\noindent\textbf{Q3 (Real-World Physical Grounding):} Does \textsc{TaskAnchor} successfully transfer to practical robotic platforms, maintaining stage consistency under multi-stage interactions and diverse historical contexts? (\emph{Sec.~\ref{sec:real_robot}})
    
\noindent\textbf{Q4 (Deployment Efficiency):} What gains in task success and additional inference costs does \textsc{TaskAnchor} introduce relative to the base VLA? (\emph{Sec.~\ref{sec:efficiency}})


\vspace{-0.1in}
\subsection{RMBench Evaluation}
\vspace{-0.08in}
\label{sec:rmbench}

\paragraph{Training and Evaluation Setup.}
RMBench~\citep{chen2026rmbench} evaluates early-evidence retrieval in bimanual manipulation, testing whether policies use previously observed object identities, locations, and poses. Performance is reported as success rate (SR) in $[0,1]$. Five $M(1)$ tasks are evaluated under the ``demo clean'' setting. In the joint setting, each \textsc{TaskAnchor} variant is trained for 250 episodes and evaluated using a single checkpoint across tasks. Each task uses 100 fixed evaluation seeds, yielding 500 closed-loop rollouts without failure filtering or seed reselection. Comparisons include DP~\citep{chi2023diffusion}, ACT~\citep{zhao2023learning},
$\pi_{0.5}$~\citep{pi05}, X-VLA~\citep{zheng2026x}, and
Mem-0~\citep{chen2026rmbench}; \textsc{TaskAnchor}-XVLA follows the same evaluation protocol. Whereas the baselines are trained separately with 50 demonstrations per task, \textsc{TaskAnchor} is evaluated under both joint multi-task training and task-specific adaptation. Joint results, denoted by $^{*}$, use a single model shared across tasks. The task-specific variant trains a separate model for each task to assess adaptation without cross-task interference.

\vspace{-0.1in}\paragraph{Evaluation Results on RMBench.}
As shown in Table~\ref{tab:rmbench_results}, \textsc{TaskAnchor} consistently improves long-horizon performance across VLA backbones and training settings. Under unified multi-task training, \textsc{TaskAnchor}-$\pi_{0.5}^{*}$ and \textsc{TaskAnchor}-XVLA$^{*}$ achieve average SR of $0.70$ and $0.66$, exceeding their corresponding pretrained backbones by $+0.55$ and $+0.54$ points, respectively. Both variants reach a success rate of $1.00$ on \emph{Rearrange}, \emph{Put Back}, and \emph{Swap Blocks}, indicating that the gains generalize across distinct VLA architectures despite joint training over all tasks. Task-specific training further increases the average SR to $0.79$ for \textsc{TaskAnchor}-$\pi_{0.5}$ and $0.70$ for \textsc{TaskAnchor}-XVLA, suggesting additional gains when cross-task interference is reduced.

The remaining failures are concentrated in \emph{Obs.\ \& Pick} and \emph{Swap T}. In \emph{Swap T}, qualitative failure cases often recover the correct object location but produce an incorrect orientation, suggesting that full-pose restoration is more challenging than spatial localization alone. In \emph{Obs.\ \& Pick}, failures are more frequently associated with target disambiguation than with grasp execution. Together, these results indicate that \textsc{TaskAnchor} substantially improves history-dependent state recovery and action selection, while fine-grained target and orientation reasoning remain challenging. Overall, the RMBench results provide system-level evidence for \textbf{Q1} and support the role of historical context in resolving action ambiguity in \textbf{Q2}.

\begin{table}[t]
\setlength{\belowcaptionskip}{6pt}
\vspace{-0.17in}
\caption{\textbf{Quantitative success rates on RMBench under the $M(1)$ setting.}
Results are rounded to two decimal places. 
$^{*}$ denotes a multi-task model trained jointly on all RMBench tasks,
whereas non-$^{*}$ methods are trained separately for each task.
The best and second-best performances in each column are highlighted in
\textbf{bold} and \underline{underlined}, respectively.}
\label{tab:rmbench_results}
\centering
\fontsize{8.5}{9.5}\selectfont
\setlength{\tabcolsep}{3pt}
\renewcommand{\arraystretch}{1}
\renewcommand{\tabularxcolumn}[1]{m{#1}}

\begin{tabularx}{\linewidth}{
    l Y{1.00} Y{1.02} Y{0.96} Y{1.16} Y{0.86} Y{1.00}
}
\toprule
\textbf{Method}
& \textbf{Obs. \& Pick}
& \textbf{Rearrange}
& \textbf{Put Back}
& \textbf{Swap Blocks}
& \textbf{Swap T}
& \textbf{Avg. SR}
\\
\midrule

DP~{\scriptsize\citep{chi2023diffusion}}
& 0.01 & 0.00 & 0.00 & 0.11 & 0.20 & 0.06
\\

ACT~{\scriptsize\citep{zhao2023learning}}
& 0.01 & 0.29 & 0.00 & 0.02 & 0.02 & 0.07
\\

$\pi_{0.5}$~{\scriptsize\citep{pi05}}
& 0.09 & 0.13 & 0.11 & 0.24 & 0.15 & 0.14
\\

X-VLA~{\scriptsize\citep{zheng2026x}}
& 0.09 & 0.13 & 0.18 & 0.16 & 0.03 & 0.12
\\

Mem-0~{\scriptsize\citep{chen2026rmbench}}
& 0.04 & \underline{0.89} & \underline{0.90}
& \underline{0.67} & 0.14 & 0.53
\\

\midrule

\rowcolor{tableblue}
\textbf{\textsc{TaskAnchor}-}$\boldsymbol{\pi_{0.5}^{*}}$
& \underline{0.16} & \textbf{1.00} & \textbf{1.00}
& \textbf{1.00} & 0.32 & \underline{0.70}
\\

\rowcolor{tableblue}
\textbf{\textsc{TaskAnchor}-XVLA}$^{*}$
& 0.04 & \textbf{1.00} & \textbf{1.00}
& \textbf{1.00} & 0.24 & 0.66
\\

\rowcolor{tableblue}
\taskanchorpibf
& \textbf{0.35} & \textbf{1.00} & \textbf{1.00}
& \textbf{1.00} & \textbf{0.62} & \textbf{0.79}
\\

\rowcolor{tableblue}
\textbf{\textsc{TaskAnchor}-XVLA}
& 0.13 & \textbf{1.00} & \textbf{1.00}
& \textbf{1.00} & \underline{0.37} & \underline{0.70}
\\

\bottomrule
\end{tabularx}\vspace{-0.1in}
\end{table}

\begin{table}[!t]
\setlength{\belowcaptionskip}{6pt}
\caption{Task Success Rate (TSR) and Cumulative Success Rate (CSR)
on RoboMemArena (\%). \textbf{Bold} and \underline{underlined}
entries indicate the best and second-best results in each column.}
\label{tab:robomemarena_results}
\centering
\fontsize{8.5}{9.5}\selectfont
\setlength{\tabcolsep}{2pt}
\renewcommand{\arraystretch}{1}
\renewcommand{\tabularxcolumn}[1]{m{#1}}

\begin{tabularx}{\linewidth}{l *{10}{Y{1}}}
\toprule
\multirow{2}{*}{\textbf{Method}}
& \multicolumn{2}{c}{\textbf{Transfer}}
& \multicolumn{2}{c}{\textbf{Occlusion}}
& \multicolumn{2}{c}{\textbf{Counting}}
& \multicolumn{2}{c}{\textbf{Sequence}}
& \multicolumn{2}{c}{\textbf{Avg.}} \\
\cmidrule(lr){2-3}
\cmidrule(lr){4-5}
\cmidrule(lr){6-7}
\cmidrule(lr){8-9}
\cmidrule(lr){10-11}
& TSR & CSR
& TSR & CSR
& TSR & CSR
& TSR & CSR
& TSR & CSR \\
\midrule

$\pi_{0.5}$~{\scriptsize\citep{pi05}}
& 20.0 & 42.8
& 12.7 & 17.2
& 14.3 & 50.9
& 60.0 & 71.6
& 21.5 & 38.7 \\

HiF-VLA~{\scriptsize\citep{lin2026hifvla}}
& 17.5 & 38.9
& 12.7 & 27.1
& 8.6 & 45.9
& 42.5 & 70.2
& 16.9 & 39.8 \\

MemoryVLA~{\scriptsize\citep{shi2026memoryvla}}
& 15.0 & 37.2
& 7.3 & 13.1
& 14.3 & 55.1
& 37.5 & 65.2
& 15.0 & 35.3 \\

MemER~{\scriptsize\citep{sridhar2026scaling}}
& 20.0 & 36.1
& 16.4 & 33.2
& 27.1 & 65.1
& 65.0 & 79.1
& 27.3 & 49.1 \\

PrediMem~{\scriptsize\citep{lei2026robomemarena}}
& \underline{22.5} & \underline{45.2}
& \underline{27.3} & \underline{38.4}
& \underline{45.7} &\underline{ 69.3}
& \textbf{72.5} & \textbf{89.5}
& \underline{38.5} & \underline{55.2} \\

\midrule
\rowcolor{tableblue}
\taskanchorpibf
& \textbf{35.0} & \textbf{47.5}
& \textbf{47.3} & \textbf{75.9}
& \textbf{51.4} & \textbf{80.0}
& \underline{70.0} & \underline{86.3}
& \textbf{50.0} & \textbf{74.2} \\
\bottomrule
\end{tabularx}
\end{table}


\subsection{RoboMemArena Evaluation}
\label{sec:robomemarena}

\paragraph{Training and Evaluation Setup.}
RoboMemArena~\citep{lei2026robomemarena} comprises 26 long-horizon
tasks averaging $1,076$ steps, with 104 of its 151 annotated
subtasks requiring history.
Transfer, occlusion, counting, and sequential-execution tasks
assess historical-state retention and execution-stage tracking.
The benchmark reports Task Success Rate (TSR), which requires
completion of all stages, and Cumulative Success Rate (CSR),
the percentage of completed stages.
Following the standard protocol, \taskanchorpi{} uses 100
demonstrations per task.
Both task-state signals are inferred from observation history
without additional inference-time annotations.
Baselines include $\pi_{0.5}$~\citep{pi05},
HiF-VLA~\citep{lin2026hifvla},
MemoryVLA~\citep{shi2026memoryvla},
MemER~\citep{sridhar2026scaling}, and
PrediMem~\citep{lei2026robomemarena}.

\vspace{-0.1in}
\paragraph{Evaluation Results on RoboMemArena.}
In Table~\ref{tab:robomemarena_results}, \taskanchorpi{} achieves
the highest average TSR ($50.0\%$) and CSR ($74.2\%$) among the
evaluated methods, exceeding $\pi_{0.5}$ by $28.5$ and $35.5$
percentage points, respectively.
Compared with PrediMem~\citep{lei2026robomemarena}, the strongest
baseline on both aggregate metrics, the corresponding gains
are $11.5$ and $19.0$ percentage points.
The largest category-level improvements over PrediMem occur
on occlusion tasks, where TSR and CSR reach $47.3\%$ and
$75.9\%$, exceeding the baseline by $20.0$ and $37.5$
percentage points.
\taskanchorpi{} also leads on both metrics in transfer and
counting, although their improvement profiles differ:
transfer gains $12.5$ points in TSR but only $2.3$ points in
CSR, whereas counting gains $5.7$ and $10.7$ points,
respectively.
Thus, the transfer advantage is more pronounced in
complete-task success, while occlusion and counting exhibit
larger gains in cumulative stage completion.
Sequence remains the only category in which \taskanchorpi{}
ranks second rather than first: its TSR of $70.0\%$ and CSR
of $86.3\%$ trail PrediMem by $2.5$ and $3.2$ percentage
points, respectively.
Overall, these results support improved long-horizon
execution (Q1), with the clearest advantage on occlusion tasks.
They are consistent with the intended benefits of
historical-state retention and execution-stage tracking (Q2),
although this comparison alone does not isolate the
contribution of either task-state signal.


\FloatBarrier
\Needspace{9\baselineskip}

\subsection{Ablation and Task-State Analysis}
\label{sec:task_state_analysis}

\paragraph{Training and Evaluation Setup.}
On RMBench $M(1)$, \taskanchorpi{}$^{*}$ is ablated by removing
the visual residual (coordinate-only) or coordinate prompting
(history-visual-only); the full model retains both.
CA-TTT depth and compressed history-token resolution are also
ablated on RMBench.
All ablations use the same training data, optimization
schedule, and evaluation seeds.


\par\Needspace{15\baselineskip}
\begin{wraptable}{r}{0.60\linewidth}
    \centering
    \setlength{\belowcaptionskip}{4pt}

    \caption{
        Component ablation of \taskanchorpi{}$^{*}$ on RMBench $M(1)$.
        Hist. Vision and Coord. denote history-conditioned visual adaptation
        and task-state coordinate prompting, respectively.
        \textbf{Bold} and \underline{underlined} entries indicate the best
        and second-best results in each column; ties receive the same formatting.
    }
    \label{tab:component_ablation}

    \fontsize{8}{9.5}\selectfont
    \setlength{\tabcolsep}{1.8pt}
    \renewcommand{\arraystretch}{1}
    \renewcommand{\tabularxcolumn}[1]{m{#1}}

    \begin{tabularx}{\linewidth}{
        @{}
        c c
        *{6}{>{\centering\arraybackslash}X}
        @{}
    }
        \toprule
        \multicolumn{2}{c}{\textbf{Components}}
        & \multicolumn{6}{c}{\textbf{Success rate}} \\
        \cmidrule(lr){1-2}
        \cmidrule(lr){3-8}

        \makecell{\textbf{Hist.}\\\textbf{Vision}}
        & \textbf{Coord.}
        & \makecell{\textbf{Obs.}\\\textbf{\& Pick}}
        & \makecell{\textbf{Rearr}\\\textbf{ange}}
        & \makecell{\textbf{Put}\\\textbf{Back}}
        & \makecell{\textbf{Swap}\\\textbf{Blocks}}
        & \makecell{\textbf{Swap}\\\textbf{T}}
        & \makecell{\textbf{Avg.}\\\textbf{SR}} \\
        \midrule

        $\times$ & $\checkmark$
        & \underline{0.13}
        & \underline{0.27}
        & \underline{0.14}
        & \underline{0.92}
        & 0.09
        & \underline{0.31} \\

        $\checkmark$ & $\times$
        & \textbf{0.16}
        & 0.17
        & \underline{0.14}
        & 0.30
        & \underline{0.25}
        & 0.20 \\

        \midrule
        \rowcolor{tableblue}
        $\checkmark$ & $\checkmark$
        & \textbf{0.16}
        & \textbf{1.00}
        & \textbf{1.00}
        & \textbf{1.00}
        & \textbf{0.32}
        & \textbf{0.70} \\

        \bottomrule
    \end{tabularx}
\end{wraptable}

\paragraph{Component Ablation.}
Using \taskanchorpi{}$^{*}$, we ablate the two task-state
conditioning pathways.
As shown in Table~\ref{tab:component_ablation}, the full model
achieves the highest average SR of $0.70$, compared with
$0.31$ for coordinate-only and $0.20$ for history-visual-only.
The full model reaches $1.00$ on Swap Block, Rearrange Block
and Put Back Block, where either single component performs
substantially worse.
The two ablated variants also show complementary behavior:
coordinate-only is stronger on Swap Blocks ($0.92$ vs.\
$0.30$), while history-visual-only performs better on Swap T
($0.25$ vs.\ $0.09$).
These results support the complementary roles of the two
pathways across RMBench tasks.

\par
\WFclear


\par\Needspace{14\baselineskip}
\begin{wraptable}{r}{0.71\linewidth}
    \centering
    \setlength{\belowcaptionskip}{6pt}

    \caption{
        \textbf{Ablation of CA-TTT depth and historical token resolution on RMBench.}
        All variants are based on $\textsc{TaskAnchor}-XVLA^{*}$.
        We vary the number of CA-TTT blocks and the spatial resolution
        of compressed historical tokens while keeping all other settings fixed.
        Success rates are measured over 100 evaluation episodes per task.
    }
    \label{tab:depth_resolution_ablation}

    \small
    \setlength{\tabcolsep}{2.5pt}

    \resizebox{\linewidth}{!}{%
        \begin{tabular}{cc|ccccc|c}
            \toprule
            \textbf{Depth}
            & \textbf{Hist. Res.}
            & \textbf{Observe}
            & \textbf{Put Back}
            & \textbf{Rearrange}
            & \textbf{Swap Blocks}
            & \textbf{Swap T}
            & \textbf{Avg. SR} \\
            \midrule

            4 & $2\times2$
            & 0.04 & 0.26 & 0.57 & 0.46 & 0.13 & 0.292 \\

            4 & $4\times4$
            & 0.03 & 1.00 & 0.78 & 0.58 & 0.13 & 0.504 \\

            8 & $2\times2$
            & 0.05 & 0.26 & 0.68 & 0.78 & 0.16 & 0.386 \\

            \rowcolor{tableblue}
            8 & \textbf{$4\times4$}
            & 0.04
            & \textbf{1.00}
            & \textbf{1.00}
            & \textbf{1.00}
            & \textbf{0.24}
            & \textbf{0.656} \\

            \bottomrule
        \end{tabular}%
    }
\end{wraptable}

\paragraph{CA-TTT Architecture Ablation.}
We ablate the temporal depth and history resolution on RMBench.
As shown in Table~\ref{tab:depth_resolution_ablation},
increasing CA-TTT depth from 4 to 8 layers improves average
SR from $0.504$ to $0.656$ with $4{\times}4$ history tokens
and from $0.292$ to $0.386$ with $2{\times}2$ tokens.
The $4{\times}4$ history resolution consistently outperforms
$2{\times}2$ compression, with the 8-layer $4{\times}4$ variant
achieving the best SR of $0.656$, which we adopt by default.
Together with component ablations and intervention, these
results support the complementary roles of the two pathways
and coordinate conditioning (\textbf{Q2}).

\par
\WFclear
\par



\par
\WFclear

\raggedbottom
\begingroup
\setlength{\parskip}{3pt}
\setlength{\intextsep}{6pt plus 1pt minus 1pt}
\setlength{\textfloatsep}{8pt plus 2pt minus 2pt}


\subsection{Performance--Efficiency Trade-off}
\label{sec:efficiency}

\paragraph{Evaluation Setup.}
We compare \taskanchorpi{} against $\pi_{0.5}$~\citep{pi05}
under identical hardware, batch size, precision, and chunk
settings, measuring inference latency per action chunk after
warm-up while excluding simulator stepping and visual I/O.


\par\addvspace{3pt}
\noindent\makebox[\linewidth][c]{%
    \begin{minipage}{0.72\linewidth}
        \centering
        \setlength{\abovecaptionskip}{0pt}
        \setlength{\belowcaptionskip}{5pt}

        \captionof{table}{
            Inference cost and average success rates on RMBench (RM)
            and RoboMemArena (RMA).
            Latency is measured per policy invocation under the native
            inference configuration of each benchmark.
            RM SR is on $[0,1]$; RMA TSR/CSR are percentages (\%).
        }
        \label{tab:efficiency}

        \fontsize{8.2}{9.2}\selectfont
        \setlength{\tabcolsep}{0.8pt}
        \renewcommand{\arraystretch}{1}
        \renewcommand{\tabularxcolumn}[1]{m{#1}}

        \begin{tabularx}{\linewidth}{
            @{}
            l
            *{2}{>{\centering\arraybackslash}X}
            c
            *{3}{>{\centering\arraybackslash}X}
            @{}
        }
            \toprule

            \textbf{Method}
            & \makecell{\textbf{RM}\\Lat.\\(ms)}
            & \makecell{\textbf{RMA}\\Lat.\\(ms)}
            & \makecell{\textbf{Added}\\\textbf{params.}\\(M)}
            & \makecell{\textbf{RM}\\SR}
            & \makecell{\textbf{RMA}\\TSR}
            & \makecell{\textbf{RMA}\\CSR} \\

            \midrule

            $\pi_{0.5}$~{\scriptsize\citep{pi05}}
            & 62.00
            & 60.83
            & --
            & 0.14
            & 21.50
            & 38.70 \\

            \rowcolor{tableblue}
            \taskanchorpibf
            & 64.08
            & 62.99
            & 26.93
            & \textbf{0.79}
            & \textbf{50.0}
            & \textbf{74.2} \\

            \bottomrule
        \end{tabularx}
    \end{minipage}%
}
\par\addvspace{3pt}

\paragraph{Performance--Efficiency Results.}
As shown in Table~\ref{tab:efficiency}, \taskanchorpi{}
improves RMBench SR by 65.0 percentage points and RoboMemArena
TSR and CSR by 28.5 and 35.5 percentage points, respectively.
It introduces 26.93M additional parameters and only
2.08\,ms and 2.16\,ms of additional inference latency per
policy invocation on RMBench and RoboMemArena, respectively,
retaining near-backbone inference efficiency (\textbf{Q4}).


\par

\subsection{Real-Robot Experiments}
\label{sec:real_robot}

\paragraph{Training and Evaluation Setup.}
We evaluate \textsc{TaskAnchor} on a physical robot platform
across three long-horizon real-world tasks that exhibit
task-state aliasing: human-cued object matching, lifting a
red block three times, and chip-bag checkout.
As illustrated in Figure~\ref{fig:real_world}, these tasks
respectively require retaining interaction-dependent cues,
tracking repeated execution stages, and maintaining state
across a multi-stage manipulation sequence.
Each task is evaluated over 25 independent trials.
Success requires completing the full sequence, including all
required intermediate state transitions, and satisfying the
final placement criteria.

\begin{figure}[!htbp]
    \centering
    \setlength{\abovecaptionskip}{4pt}
    \setlength{\belowcaptionskip}{0pt}
    \includegraphics[width=0.95\linewidth]{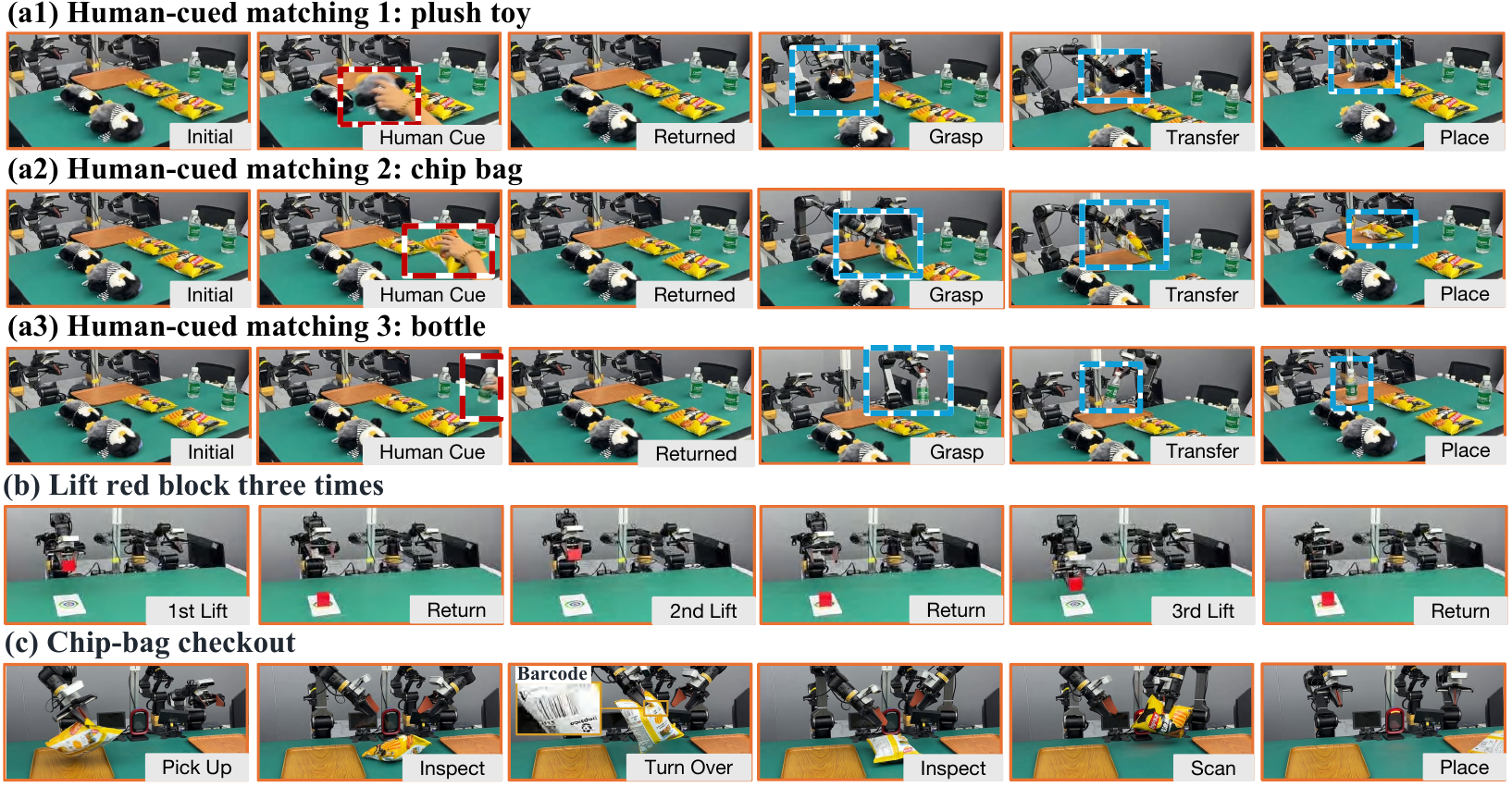}

    \caption{
        \textbf{Real-world tasks for task-state-conditioned manipulation.}
        \textbf{(a) Human-cued matching:}
        after a human removes and returns a cue object, the robot
        identifies and grasps the corresponding object from the
        bottom row, demonstrating the use of external interaction
        history as temporal context.
        \textbf{(b) Lift red block three times:}
        repeated lift--return cycles three times require
        maintaining execution stage and terminating after the
        third repetition.
        \textbf{(c) Chip-bag checkout:}
        the robot performs Chip-bag inspection, barcode
        orientation adjustment, scanning, and final placement,
        requiring consistent multi-stage state tracking.
        All panels show cropped third-person RGB frames ordered
        from left to right.
        Colored boxes indicate human cues and corresponding targets.
        Further details of the real-robot experiments are provided
        in Appendix~\ref{app:real_evaluation}.
    }
    \label{fig:real_world}
\end{figure}


\par\addvspace{3pt}
\noindent
\begin{minipage}{\linewidth}

    \begin{minipage}[t]{0.47\linewidth}
        \vspace{0pt}

        \paragraph{Evaluation Results on Real-Robot Tasks.}
        As shown in Table~\ref{tab:real_robot_results},
        \textsc{TaskAnchor} achieves $100\%$ success
        ($25 / 25$) on each task, compared with $0 / 25$,
        $1 / 25$, and $4 / 25$ for
        X-VLA~\citep{zheng2026x} on human-cued matching,
        repeated lifting, and chip-bag checkout, respectively.
    \end{minipage}\hfill
    %
    \begin{minipage}[t]{0.50\linewidth}
        \vspace{0pt}
        \centering
        \setlength{\abovecaptionskip}{0pt}
        \setlength{\belowcaptionskip}{6pt}

        \captionof{table}{
            \textbf{Real-world evaluation results.}
            Success rates are measured over 25 independent trials.
        }
        \label{tab:real_robot_results}

        \fontsize{8.2}{9.2}\selectfont
        \setlength{\tabcolsep}{2pt}
        \renewcommand{\arraystretch}{1.05}
        \renewcommand{\tabularxcolumn}[1]{m{#1}}

        \begin{tabularx}{\linewidth}{
            @{}
            >{\raggedright\arraybackslash}X
            c
            c
            @{}
        }
            \toprule

            Task
            & {\scriptsize X-VLA}
            & \makecell{
                {\scriptsize\textsc{TaskAnchor}-}\\[-1pt]
                {\scriptsize XVLA}
            } \\

            \midrule

            Human-cued matching
            & 0 / 25
            & 25 / 25 \\

            Lift red block three times
            & 1 / 25
            & 25 / 25 \\

            Chip-bag checkout
            & 4 / 25
            & 25 / 25 \\

            \bottomrule
        \end{tabularx}
    \end{minipage}

\end{minipage}
\par\addvspace{4pt}


\noindent
Observed baseline failures include premature manipulation
before the human removal-and-return cue is complete,
excessive repetitions despite successful individual lifts,
and inconsistent transitions among inspection, reorientation,
scanning, and placement.
These failures highlight the distinction between executing
local actions and selecting them at the appropriate subtask
stage.
The performance contrast shows more reliable stage-dependent
execution and history-dependent action selection with
\textsc{TaskAnchor}.
In particular, successful human-cued matching provides
evidence that temporal context can extend beyond robot
execution history to external human interactions that
determine subsequent target selection.
Additional execution sequences and baseline failure cases
are provided in Appendix~\ref{app:failure_analysis}.

\par
\endgroup


\vspace{-0.12in}
\section{Conclusion}
\vspace{-0.1in}
\label{sec:conclusion}

We present \textsc{TaskAnchor}, showing that lightweight task-state adaptation enables reactive VLAs to handle complex long-horizon manipulation with modest latency overhead. Beyond empirical gains, our findings highlight a shift from standard \emph{language-guided} policies to \emph{context-guided} VLAs. By combining history-conditioned visual features with an explicit task-state coordinate, \textsc{TaskAnchor} provides a richer conditioning space for fine-grained, state-consistent action generation without an explicit planning loop.

Our results suggest that lightweight task-state conditioning helps pretrained VLAs reuse their manipulation skills for history-dependent action selection. This points toward a broader direction: learning transferable interfaces that connect historical context, including human demonstration and robot execution history, to the control capabilities of pretrained policies. 
This formulation offers a path toward scalable embodied agents that adapt their behavior to diverse evolving task contexts during long-horizon manipulation.



\newpage
\bibliography{iclr2027_conference}
\bibliographystyle{iclr2027_conference}

\newpage
\appendix

\section{Implementation Details}
\label{app:implementation}

\subsection{Base Models and Training Configurations}
\label{app:base_models}

We evaluate \textsc{TaskAnchor} on two representative pretrained vision-language-action models, X-VLA~\citep{zheng2026x} and $\pi_{0.5}$~\citep{pi05}. \textsc{TaskAnchor} is learned during standard VLA post-training through supervised fine-tuning (SFT) and introduces only a lightweight adapter, while preserving the original backbone architecture, vision-language fusion mechanism, and action generation interface.

We initialize X-VLA experiments from the publicly available X-VLA-PT checkpoint. X-VLA adopts a DaViT~\citep{ding2022davit} visual backbone that produces 2048-dimensional visual features. During training, the DaViT visual encoder remains frozen, while \textsc{TaskAnchor} and enabled downstream adaptation parameters are optimized on top of the pretrained representation. The enabled downstream components include the image projection layer, language-side adaptation modules, soft prompt parameters, action core, and action head.

For $\pi_{0.5}$~\citep{pi05} experiments, we initialize from the official $\pi_{0.5}$ model. Unlike X-VLA~\citep{zheng2026x}, $\pi_{0.5}$ adopts the native SigLIP~\citep{tschannen2025siglip} visual representation space with 1152-dimensional visual features. The cached visual features remain frozen in the \textsc{TaskAnchor} variant, while \textsc{TaskAnchor} and selected downstream adaptation parameters are optimized through standard SFT. The official $\pi_{0.5}$ full-SFT baseline follows the native supervised fine-tuning procedure.

Both \textsc{TaskAnchor} variants adopt the same optimization principle. During the first 5K optimizer updates, only \textsc{TaskAnchor} parameters are optimized to stabilize the learning of the history-conditioned visual residual and task-state coordinate. After this adapter warm-up stage, enabled downstream adaptation components are jointly optimized with \textsc{TaskAnchor}.

The \textsc{TaskAnchor}-XVLA model is trained for 55K optimizer updates. For \textsc{TaskAnchor}-$\pi_{0.5}$, the RMBench experiments use 35K optimizer updates, whereas the RoboMemArena results in Table~\ref{tab:robomemarena_results} are evaluated using the 50K-step checkpoint. All experiments are conducted on NVIDIA A800 GPUs using BF16 precision. Each experiment uses eight GPUs with 32 samples per GPU, resulting in a global batch size of 256. Trainable parameters are optimized using AdamW with $\beta_1=0.9$ and $\beta_2=0.95$, together with gradient norm clipping of 1.0.

\begin{table}[htbp]
\centering
\setlength{\belowcaptionskip}{6pt}
\caption{Summary of backbone and training configurations.}
\label{tab:training_config}
\small
\setlength{\tabcolsep}{6pt}
\begin{tabular}{lcc}
\toprule
 & \textsc{TaskAnchor}-XVLA & \taskanchorpi \\
\midrule
Initialization
& X-VLA-PT
& $\pi_{0.5}$-PT \\

Visual representation
& DaViT
& SigLIP \\

Feature dimension
& 2048
& 1152 \\

Visual encoder
& Frozen
& Frozen cached feature \\

Training method
& SFT
& SFT \\

\textsc{TaskAnchor} warm-up
& 5K steps
& 5K steps \\

Total training steps
& 55K
& 35K (RM) / 50K (RMA) \\

Hardware
& 8$\times$A800
& 8$\times$A800 \\

Precision
& BF16
& BF16 \\

Global batch size
& 256
& 256 \\

\bottomrule
\end{tabular}
\end{table}

\subsection{\textsc{TaskAnchor} Architecture Configuration}
\label{app:taskanchor_architecture}

\textsc{TaskAnchor} is implemented as a lightweight adapter between the visual
encoder and the original multimodal fusion module of a pretrained VLA (Figure~\ref{fig:pipeline}(a)). Given
current backbone visual tokens
$X_t\in\mathbb{R}^{N_t\times d_b}$ and historical tokens
$H_t\in\mathbb{R}^{N_h\times d_b}$, where $d_b$ denotes the
backbone-specific visual feature dimension, \textsc{TaskAnchor} first projects
both representations into a shared latent space:

\begin{equation}
\bar X_t=P_{\mathrm{in}}(X_t),\qquad
\bar M_t=P_{\mathrm{in}}(H_t),
\qquad
P_{\mathrm{in}}:\mathbb{R}^{d_b}\rightarrow\mathbb{R}^{512}.
\end{equation}

The projected tokens $\bar X_t$ and $\bar M_t$ are processed by an eight-layer CA-TTT module operating entirely in the 512-dimensional latent space. A backbone-specific output projection $P_{\mathrm{out}}:\mathbb{R}^{512}\rightarrow\mathbb{R}^{d_b}$ restores the original feature dimension and produces the history-conditioned visual residual. A separate task-state coordinate head operates on the
coordinate token representation. The adapted visual representation and task-state condition are then passed through the original VLA fusion and action-generation modules. \textsc{TaskAnchor} does not modify the pretrained VLA architecture or introduce an additional action decoder.

\paragraph{Temporal adapter.}
The temporal adapter consists of 8 pre-normalized CA-TTT blocks with hidden dimension 512, 8 attention heads, head dimension 64, and MLP ratio 4. Each block receives a query sequence consisting of one learnable task-state coordinate token followed by current visual tokens, while historical visual tokens serve as temporal memory.

Each CA-TTT block contains a temporal adaptation module followed by a pre-normalized MLP residual branch (Figure~\ref{fig:pipeline}(b)). All attention, normalization, and feed-forward operations are performed in the 512-dimensional latent space. The MLP hidden dimension is $4\times512=2048$. We use element-wise dropout with rate 0.1, and set the LayerNorm~\citep{ba2016layer} epsilon to $10^{-6}$.

The fast weights used by CA-TTT are temporary tensors constructed during each forward pass. During inference, they are recomputed from the current history buffer at every replanning step and discarded after the action chunk is generated. They are not stored in checkpoints, propagated across replanning calls, or updated by an inference-time optimizer. Only the external history buffer is maintained online. During training, the same temporal adaptation procedure is included in the differentiable forward computation.

\paragraph{Backbone-specific interfaces.}
\textsc{TaskAnchor} shares the same 512-dimensional temporal adaptation module across different pretrained VLAs while using backbone-specific feature projections. For X-VLA, the adapter consumes 2048-dimensional DaViT~\citep{ding2022davit} features and applies a $2048\rightarrow512\rightarrow2048$ projection path. For $\pi_{0.5}$, the adapter operates on 1152-dimensional SigLIP~\citep{tschannen2025siglip} features with a $1152\rightarrow512\rightarrow1152$ projection path. After adaptation, the original token ordering of each backbone is restored before the representation is passed to the native multimodal fusion module.

\begin{table}[t]
\centering
\caption{Backbone-specific \textsc{TaskAnchor} interfaces.}
\label{tab:taskanchor_backbone_config}
\small
\setlength{\tabcolsep}{5pt}
\begin{tabular}{lcc}
\toprule
Configuration & X-VLA & $\pi_{0.5}$ \\
\midrule
Visual encoder & DaViT & SigLIP \\
Feature dimension & 2048 & 1152 \\
Projection path & $2048\rightarrow512\rightarrow2048$ & $1152\rightarrow512\rightarrow1152$ \\
Current tokens/view & 50 & 256 \\
\bottomrule
\end{tabular}
\end{table}

\paragraph{Visual residual injection.}
After the final CA-TTT block, the adapted latent visual tokens are
normalized and projected back to the backbone feature dimension:
\begin{equation}
\Delta X_t
=
P_{\mathrm{out}}
\left(
\mathrm{LN}\left(\bar X_t^{(L)}\right)
\right),
\end{equation}
where $\bar X_t^{(L)}$ denotes the visual-token representation after the
final CA-TTT block. The history-conditioned visual representation is then
obtained through residual injection:
\begin{equation}
\widetilde X_t
=
X_t+\Delta X_t.
\end{equation}

\paragraph{Parameter composition.}
The number of \textsc{TaskAnchor} parameters depends on the backbone feature dimension because the input and output projection layers are backbone-specific. For the X-VLA ~\citep{zheng2026x} instantiation, \textsc{TaskAnchor} contains exactly $27,850,242$ trainable parameters, which is reported as approximately 28M. This count includes the input and output projections, 8 CA-TTT blocks, normalization layers, task-state coordinate token and prediction head, and all adapter-specific parameters. It excludes the frozen DaViT encoder and all pre-existing X-VLA parameters.

For $\pi_{0.5}$~\citep{pi05}, the $1152\rightarrow512\rightarrow1152$ interface results in $26,931,841$ trainable parameters. The difference originates from the backbone-specific projection layers.

\begin{table}[t]
\centering
\caption{Trainable parameter composition of TaskAnchor.}
\label{tab:taskanchor_parameter}
\small
\begin{tabular}{lcc}
\toprule
Component & X-VLA & $\pi_{0.5}$ \\
\midrule
8 $\times$ CA-TTT block & 25.743M & 25.743M \\
Input projection & 1.049M & 0.590M \\
Output projection & 1.051M & 0.591M \\
Coordinate branch & 0.007M & 0.007M \\
Total & 27.850M & 26.932M \\
\bottomrule
\end{tabular}
\end{table}

\subsection{Inference Configuration}
\label{app:inference_configuration}

During inference, \textsc{TaskAnchor} preserves the original execution interface of pretrained VLAs while maintaining an additional online history buffer for task-state estimation. At each replanning step, the current observation is first encoded by the frozen visual encoder. \textsc{TaskAnchor} then processes the current visual representation together with the accumulated history to produce the history-conditioned visual residual and predicted task-state coordinate. The adapted visual representation and task-state condition are subsequently passed to the original VLA fusion module and action decoder to generate the next action chunk. No architectural modification is introduced to the pretrained VLA during inference.

\paragraph{Online history update.}
\textsc{TaskAnchor} maintains an external visual-history buffer during execution. After each replanning step, newly observed visual features are inserted into the history buffer according to the backbone-specific sampling strategy described in Appendix~\ref{app:history_memory}. The history buffer stores visual representations rather than raw images, and no additional recurrent state is introduced into the pretrained VLA. The temporary fast weights used by CA-TTT are reconstructed from the current history representation at each replanning step and discarded after the action chunk is generated. They are not stored in checkpoints, propagated across replanning calls, or updated by an inference-time optimizer. Therefore, the only persistent execution memory is the external history buffer.

\paragraph{Task-state conditioning during inference.}
During inference, the task-state coordinate is predicted online from the current observation and accumulated history. The predicted coordinate is formatted into the corresponding task condition and injected through the native language-conditioning interface of the pretrained VLA. The original tokenizer, text embedding layers, and action decoder are reused without modification. Since the adapted visual representation is already computed, task-state conditioning does not require an additional visual encoder forward pass.

\paragraph{Action execution and replanning protocol.}
\textsc{TaskAnchor} follows the native action-chunk execution protocol of each benchmark. We distinguish the prediction horizon, execution horizon, and replanning horizon. The prediction horizon denotes the number of future actions produced by one policy invocation, whereas the execution and replanning horizons specify how many control steps are executed before a new observation is acquired and the policy is invoked again.

For RMBench \citep{chen2026rmbench}, both X-VLA~\citep{zheng2026x} and $\pi_{0.5}$ use a $30/30/30$ prediction--execution--replanning configuration. The policy predicts 30 future actions, executes the complete chunk, and replans from a newly acquired observation every 30 control steps. For RoboMemArena~\citep{lei2026robomemarena},  $\pi_{0.5}$ follow the benchmark protocol with a $20/10/10$ configuration. The policy predicts 20 future actions but executes only the first 10 before acquiring a new observation and replanning; the remaining actions from the previous prediction are discarded.

\begin{table}[t]
\centering
\caption{Inference configurations used in different evaluations.}
\label{tab:inference_configuration}
\small
\begin{tabular}{lcc}
\toprule
Evaluation & X-VLA & $\pi_{0.5}$ \\
\midrule
RMBench & $30/30/30$ & $30/30/30$ \\
RoboMemArena & $20/10/10$ & $20/10/10$ \\
\bottomrule
\end{tabular}
\end{table}

\section{Additional Method Details}

\subsection{History Memory Construction}
\label{app:history_memory}

\textsc{TaskAnchor} maintains an external visual-history buffer (Figure~\ref{fig:pipeline}(b)) that is updated at the native replanning frequency of each benchmark. At every replanning step, the newly acquired observation is passed through the pretrained visual encoder. Its visual tokens serve as the current observation for the native VLA pathway and as the query input to CA-TTT, while previously stored tokens form the history memory. After the current policy forward pass, the encoded observation is appended to the history buffer and becomes available to subsequent replanning steps. This ordering prevents the current observation from being included in its own history context.

Accordingly, both observation encoding and history insertion occur every 30 control steps on RMBench\citep{chen2026rmbench} and every 10 control steps on RoboMemArena~\citep{lei2026robomemarena}. Observations encountered during execution between two replanning points are used only by the benchmark's native low-level action execution process and are not separately encoded or inserted into the \textsc{TaskAnchor} history buffer.

For X-VLA~\citep{zheng2026x}, the visual encoder is based on DaViT~\citep{ding2022davit}. Each camera view produces one global token and 49 spatial tokens, yielding 50 visual tokens per current observation. For historical observations, the original $7\times7$ spatial feature grid is bilinearly reduced to a $4\times4$ grid, producing 16 history tokens per frame and camera view. The default X-VLA \textsc{TaskAnchor} implementation uses the compressed main-camera tokens as the CA-TTT history memory, while the current wrist-camera tokens continue through the original X-VLA visual-language fusion pathway without temporal residual modification.

For $\pi_{0.5}$~\citep{pi0}, \textsc{TaskAnchor} operates on the native SigLIP~\citep{tschannen2025siglip} visual representation, where each current camera view contains 256 visual tokens. Historical observations are spatially compressed to a $4\times4$ grid, yielding 16 tokens per view and 48 tokens across the three camera views. Similar to X-VLA, the temporal adaptation module uses the compressed main-camera historical tokens as the CA-TTT memory while preserving the pretrained $\pi_{0.5}$ multi-camera token organization for the current observation.

\subsection{CA-TTT Computation Details}
\label{app:catt_details}

The temporal adaptation module in \textsc{TaskAnchor} implements the
context-adaptive test-time-training (CA-TTT) mechanism described in
Sec.~3.2. Unlike conventional cross-attention, which directly aggregates
historical tokens through an attention matrix, CA-TTT first constructs
temporary fast weights from historical representations and then reads these
weights using the current query stream.

For completeness, we instantiate the layer-specific key and value projections
$f_k^{(m)}$ and $f_v^{(m)}$ introduced in Sec.~3.2, together with the
query projection used internally by each CA-TTT block. Given the input query
stream $Z_t^{(m-1)}$ to the $m$-th block and the projected historical memory
$\bar{M}_t$, we compute

\begin{equation}
\bar{Q}_t^{(m)}
=
\mathrm{LN}(Z_t^{(m-1)})W_Q^{(m)},
\qquad
K_t^{(m)}
=
\mathrm{LN}(\bar{M}_t)W_K^{(m)},
\qquad
V_t^{(m)}
=
\mathrm{LN}(\bar{M}_t)W_V^{(m)}.
\end{equation}

Here, $K_t^{(m)}$ and $V_t^{(m)}$ instantiate the layer-specific mappings
$f_k^{(m)}(M_t)$ and $f_v^{(m)}(M_t)$ from the main text, while
$\bar{Q}_t^{(m)}$ denotes the projected query representation used internally
within the $m$-th CA-TTT block. The query stream contains the learnable
task-state coordinate token together with the current visual tokens, whereas
the historical tokens form the memory sequence.

Within each CA-TTT block, the inner model
$g(\cdot;\omega^{(m)})$ introduced in Sec.~3.2 is implemented as a lightweight
gated MLP parameterized by two learned slow matrices:

\begin{equation}
\omega^{(m)}
=
\left\{
W_1^{(m)},W_2^{(m)}
\right\}.
\end{equation}

These slow weights are adapted during the forward pass using the
history-dependent self-supervised objective. Specifically, the temporary fast
weights are constructed as

\begin{equation}
\widehat{W}_{i,t}^{(m)}
=
W_i^{(m)}
-
\eta_{\mathrm{ttt}}
\frac{
\nabla_{W_i^{(m)}}
\mathcal{L}_{\mathrm{TTT},t}^{(m)}
\left(
K_t^{(m)},V_t^{(m)}
\right)
}{
\left\|
\nabla_{W_i^{(m)}}
\mathcal{L}_{\mathrm{TTT},t}^{(m)}
\left(
K_t^{(m)},V_t^{(m)}
\right)
\right\|_2+1
},
\qquad i\in\{1,2\}.
\end{equation}

This is the normalized implementation of the fast-weight update
$\widehat{\omega}_t^{(m)}
=
\omega^{(m)}
-
\eta_{\mathrm{ttt}}
\nabla_{\omega}
\mathcal{L}_{\mathrm{TTT},t}^{(m)}$
written in simplified form in Eq.~(4).
The normalization is used for stable fast-weight adaptation.
The resulting fast weights are temporary variables conditioned on the current
history and are discarded after the forward pass; they are not stored as
persistent model parameters.

The adapted inner MLP is then queried by the current representation through
the gated transformation

\begin{equation}
R_t^{(m)}
=
\left(
Q_t^{(m)}\widehat{W}_{1,t}^{(m)}
\right)
\odot
\mathrm{SiLU}\!\left(
Q_t^{(m)}\widehat{W}_{2,t}^{(m)}
\right).
\end{equation}

We emphasize that this history-adapted inner MLP is distinct from the
feed-forward MLP that follows the CA-TTT temporal adaptation module within
each block. The former constitutes the history-conditioned fast-weight
mechanism and operates with the temporary parameters
$\widehat{\omega}_t^{(m)}$ constructed above. The latter is a standard
pre-normalized feed-forward residual branch with hidden dimension 2048
($4\times$ the 512-dimensional adapter width), whose parameters are ordinary
offline-learned slow weights and are not adapted from the history at
inference time.

During training, fast-weight construction remains differentiable, allowing the
slow CA-TTT parameters and the remaining \textsc{TaskAnchor} parameters to be
optimized by the outer objective. During inference, the fast weights are
reconstructed independently from the current history buffer at every
replanning step and discarded after the corresponding action chunk is
generated. No persistent parameter update is performed during deployment.

\subsection{Task-state Coordinate Construction}
\label{app:task_state_coordinate}

\textsc{TaskAnchor} represents the semantic execution stage of a manipulation task using a continuous task-state coordinate. Unlike spatial coordinates, the task-state coordinate encodes the current semantic stage within the task execution process described in Section~\ref{sec:task_coordinate}. 

\paragraph{Coordinate prediction head.}
During \textsc{TaskAnchor} computation, a learnable task-state coordinate token is prepended to the current visual query sequence and processed jointly with the visual tokens through the CA-TTT module. After the final temporal adaptation layer, the hidden state of the coordinate token is used to predict the task-state coordinate.

Specifically, given the final coordinate token representation
$h_t^p\in\mathbb{R}^{512}$, the prediction head is implemented as:

\begin{equation}
p_t=
\sigma
(
W_p\operatorname{LN}(h_t^p)+b_p
),
\qquad
p_t\in(0,1),
\end{equation}

where $\sigma(\cdot)$ is the sigmoid activation. The coordinate prediction head is independent of the visual residual projection and shares only the temporal adaptation backbone. Therefore, the coordinate branch and visual residual branch provide complementary outputs from the same history-conditioned representation.

\paragraph{Task-state conditioning.}
During training, the ground-truth task-state coordinate $p_t^{\star}$ is used to construct the task condition. During inference, the predicted coordinate $p_t$ replaces the ground-truth value. The coordinate is converted into a textual condition through the native language-conditioning interface of the pretrained VLA:

\begin{equation}
\mathrm{Tokenize}\!\left(
\texttt{<coords>}~\rho(p_t)~\texttt{</coords>}
\oplus
\texttt{<task>}~\ell~\texttt{</task>}
\right),
\end{equation}

where $\rho(\cdot)$ converts the scalar coordinate into its normalized decimal string representation, and $\oplus$ denotes textual concatenation.

The resulting task condition is processed by the original tokenizer and language embedding layers of the pretrained VLA. No additional language encoder or task-specific action head is introduced. The same action generation pathway is therefore preserved between the original VLA and TaskAnchor.

\section{Experimental Setup}
\label{app:experimental_setup}

This section provides additional details of the experimental settings, including benchmark configurations, real-world robotic platforms, data collection procedures, and evaluation protocols.

\subsection{Benchmark Settings}
\label{app:benchmark_settings}

We evaluate \textsc{TaskAnchor} on RMBench\citep{chen2026rmbench} and RoboMemArena~\citep{lei2026robomemarena}, which are designed to assess long-horizon manipulation under history-dependent decision scenarios. All experiments follow the original observation interfaces, action representations, and evaluation protocols of the corresponding benchmarks.

RMBench\citep{chen2026rmbench} evaluates early-evidence retrieval in bimanual
manipulation, whereas RoboMemArena~\citep{lei2026robomemarena} contains longer
tasks involving transfer, occlusion, counting, and sequential execution.
RMBench reports success rate (SR) within $[0,1]$; RoboMemArena reports Task
Success Rate (TSR), requiring completion of all stages, and Cumulative Success
Rate (CSR), measuring the percentage of completed stages.

\paragraph{RMBench.}
RMBench\citep{chen2026rmbench} evaluates whether a manipulation policy can correctly condition its actions on previous interaction history rather than relying only on the current observation. The benchmark contains long-horizon manipulation tasks where visually similar observations may correspond to different execution stages or require different future actions depending on previous interactions. Each episode starts from a predefined initial configuration, and the policy receives RGB observations together with task instructions. The policy generates action chunks according to its native execution protocol. A trial is considered successful only when the complete task objective is achieved without external intervention.

For RMBench evaluation, both X-VLA~\citep{zheng2026x} and $\pi_{0.5}$ follow the $30/30/30$ prediction-execution-replanning protocol. The policy predicts 30 future actions and executes the complete action chunk before obtaining a new observation and replanning.

\paragraph{RoboMemArena.}
RoboMemArena~\citep{lei2026robomemarena} evaluates memory-dependent manipulation in longer-horizon and real-world interaction scenarios. The benchmark focuses on whether a robot can preserve and utilize task-relevant information after previous interactions that are no longer directly observable from the current view.

We preserve the benchmark-provided observation interface, action representation, and evaluation protocol. For RoboMemArena experiments,  $\pi_{0.5}$ follow the $20/10/10$ prediction-execution-replanning protocol, where the policy predicts 20 future actions and executes the first 10 actions before replanning.

\subsection{Real-world Platform and Data Collection}
\label{app:real_platform}

\paragraph{Robot platform.}
We conduct real-world experiments on a dual-arm manipulation platform comprising two 6-DoF PiPER robotic arms, each equipped with dual parallel-jaw grippers. The vision system contains one table-mounted head camera and two wrist-mounted cameras, all of which are Orbbec DaBai binocular depth sensors. The complete hardware configuration is illustrated in Appendix Figure~\ref{fig:real_world_setup}. The platform follows the standard multi-view observation interface used by pretrained VLAs, including synchronized RGB observations from the head and wrist cameras.

\begin{figure}[t]
    \centering
    \includegraphics[angle=-90,width=0.6\linewidth]{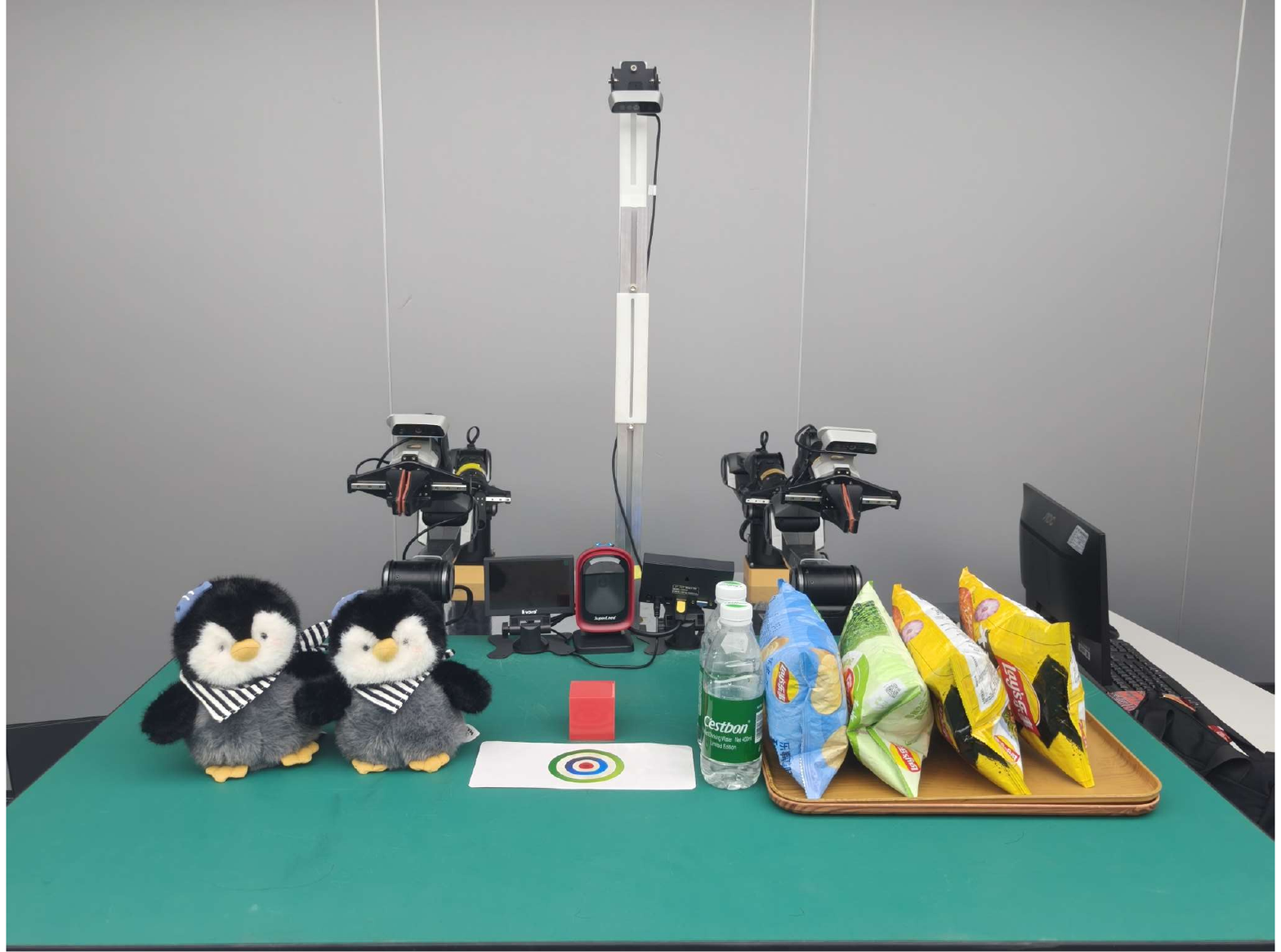}
    \caption{\textbf{Real-world robotic platform.} The platform consists of two 6-DoF PiPER robotic arms with parallel-jaw grippers, one table-mounted head camera, and two wrist-mounted cameras. All cameras use Orbbec DaBai binocular depth sensors. The setup provides synchronized multi-view RGB observations for long-horizon manipulation experiments.}
    \label{fig:real_world_setup}
\end{figure}

\paragraph{Demonstration collection.}
We collect real-world demonstrations using the same observation interface as the deployed policy. Each demonstration records synchronized multi-view RGB observations, robot states, and action trajectories. The collected trajectories contain long-horizon manipulation sequences involving multiple execution stages, requiring the robot to preserve task-relevant information across intermediate interactions.

The collected real-world tasks include three representative categories: long-horizon object manipulation, visual recognition and scanning, and human-context memory tasks. These tasks are designed to evaluate whether a policy can correctly condition actions on execution history under visually ambiguous observations.

\subsection{Real-world Evaluation Setup}
\label{app:real_evaluation}

We evaluate \textsc{TaskAnchor} on three real-world manipulation scenarios: long-horizon counting manipulation, supermarket QR scanning, and human-context object retrieval.

\textbf{Human-cued object matching.}
This task evaluates history-dependent branch selection under external intervention. Following the instruction ``After the human removes an object from the table and puts it back, pick up the corresponding object in the bottom row.'', a human removes one object from three candidates and returns it to the original scene. The robot must identify the corresponding object from the bottom row, transfer it to the tray, and retract. Although the final scene is visually similar across trials, the correct action depends on the hidden interaction history introduced by the human. This task evaluates whether \textsc{TaskAnchor} can extend beyond the robot's own execution history and leverage human demonstrations as temporal context for action selection.

\textbf{Lift red block three times.}
This task evaluates execution-stage identification in repetitive manipulation. The robot is instructed to lift a red block three times and terminate after completing all repetitions. Each lift--return cycle leads to a visually similar configuration, making the current observation insufficient to determine whether the required number of repetitions has been reached. Successful execution therefore requires maintaining an internal task state that tracks the current repetition stage.

\textbf{Chip-bag checkout.}
This task evaluates multi-stage state tracking during object interaction. The robot first picks up a chip bag from the tray, moves it to the scanning area, adjusts the barcode orientation, performs scanning, and places the object onto the target tray after successful recognition. A successful scan is determined by the green indicator displayed on the tabletop reader. This task requires coordinating multiple intermediate states, including object manipulation, orientation adjustment, scanning verification, and final placement.

All real-world evaluations are performed without additional online fine-tuning. \textsc{TaskAnchor} inference follows the deployment procedure described in Appendix~\ref{app:inference_configuration}, including online history accumulation and native VLA action generation.

\providecommand{\RRResultsFigureRoot}{figures}

\section{Additional Analysis}
\label{app：add_analysis}
\subsection{Comparison with Related Adaptation Strategies}
\label{app:related_adaptation}

\textsc{TaskAnchor} is related to recent approaches that enhance pretrained
vision-language-action models with historical information and lightweight
memory mechanisms~\citep{koo2026hamlet,shi2026memoryvla,jiang2026robottt}. These methods demonstrate that long-horizon
manipulation benefits from providing policies with information beyond the
current observation. For example, memory-based VLA adaptation methods
compress previous observations into compact representations and aggregate
them to provide temporally informed conditions for action prediction.

However, \textsc{TaskAnchor} differs from existing history-aware adaptation
strategies in how historical information is represented and injected.
Rather than treating history as an additional context source only,
\textsc{TaskAnchor} explicitly grounds task state by transforming history into
two complementary conditions: a history-conditioned visual residual and a
task-state coordinate. The visual residual preserves decision-relevant
interaction evidence, while the coordinate provides an explicit semantic
stage condition for action selection.

This design provides several advantages. First, \textsc{TaskAnchor} separates
execution-state estimation from action generation while preserving the
original VLA policy. The pretrained action decoder and fusion pathway
remain unchanged, and task-state information is injected through the
native visual and language interfaces. Second, \textsc{TaskAnchor} decouples
temporal adaptation from backbone-specific visual dimensions by performing
history reasoning in a shared 512-dimensional latent space. Therefore, the
same temporal adaptation module can be applied to VLAs with different
visual encoders, such as X-VLA~\citep{zheng2026x} and $\pi_{0.5}$~\citep{pi05}, through lightweight
interface projections. Third, \textsc{TaskAnchor} maintains flexible history
accumulation through its linear-time temporal adaptation mechanism rather
than relying on directly extending the backbone context window.

Overall, \textsc{TaskAnchor} complements existing memory-based VLA adaptation by
not only preserving historical information but also extracting the coordinate required for selecting appropriate behaviors under
visually similar observations.

\subsection{Additional Real-Robot Results and Failure Analysis}
\label{app:failure_analysis}

We provide temporally ordered execution sequences for the three real-world tasks introduced in Sec.~4.5, together with qualitative analysis of the corresponding X-VLA~\citep{zheng2026x} failure patterns. These examples complement the quantitative results in Table~6 by illustrating how \textsc{TaskAnchor} and the reactive X-VLA baseline behave across history-dependent task-state transitions. Each visualization contains 28 selected frames ordered from left to right and top to bottom. The upper-left number denotes the displayed frame index, and the upper-right timestamp indicates the corresponding time in the source clip or extracted segment. Sampling intervals are nonuniform; the sequences summarize the recorded executions rather than showing every video frame. Dark frame borders highlight representative cues or execution milestones in \textsc{TaskAnchor} rollouts, whereas red frame borders highlight repeated attempts without stage progression or visible cue--action mismatches in the X-VLA comparisons.

Across the three tasks, a common qualitative pattern emerges. X-VLA can often execute local manipulation behaviors such as grasping, lifting, moving, and placing, but its behavior becomes unreliable when selecting the appropriate behavior depends on execution context that is not recoverable from the current observation alone. The examples below illustrate three complementary forms of history-dependent ambiguity: repetition-stage identification, multi-stage transition tracking, and external-cue-dependent action-branch selection.

\subsubsection{Lift Red Block Three Times}
\label{app:real_robot_long_horizon}

The repeated-lifting task evaluates execution-stage identification under recurrent visual states. The robot must lift and return the red block three times and terminate after the third repetition. Because each return restores a similar visible configuration, observations from different repetition counts can be visually similar even though they correspond to different execution stages.

\begin{figure}[p]
    \centering
    \includegraphics[
        width=\linewidth,
        height=0.84\textheight,
        keepaspectratio,
        trim=10bp 38bp 10bp 50bp,
        clip
    ]{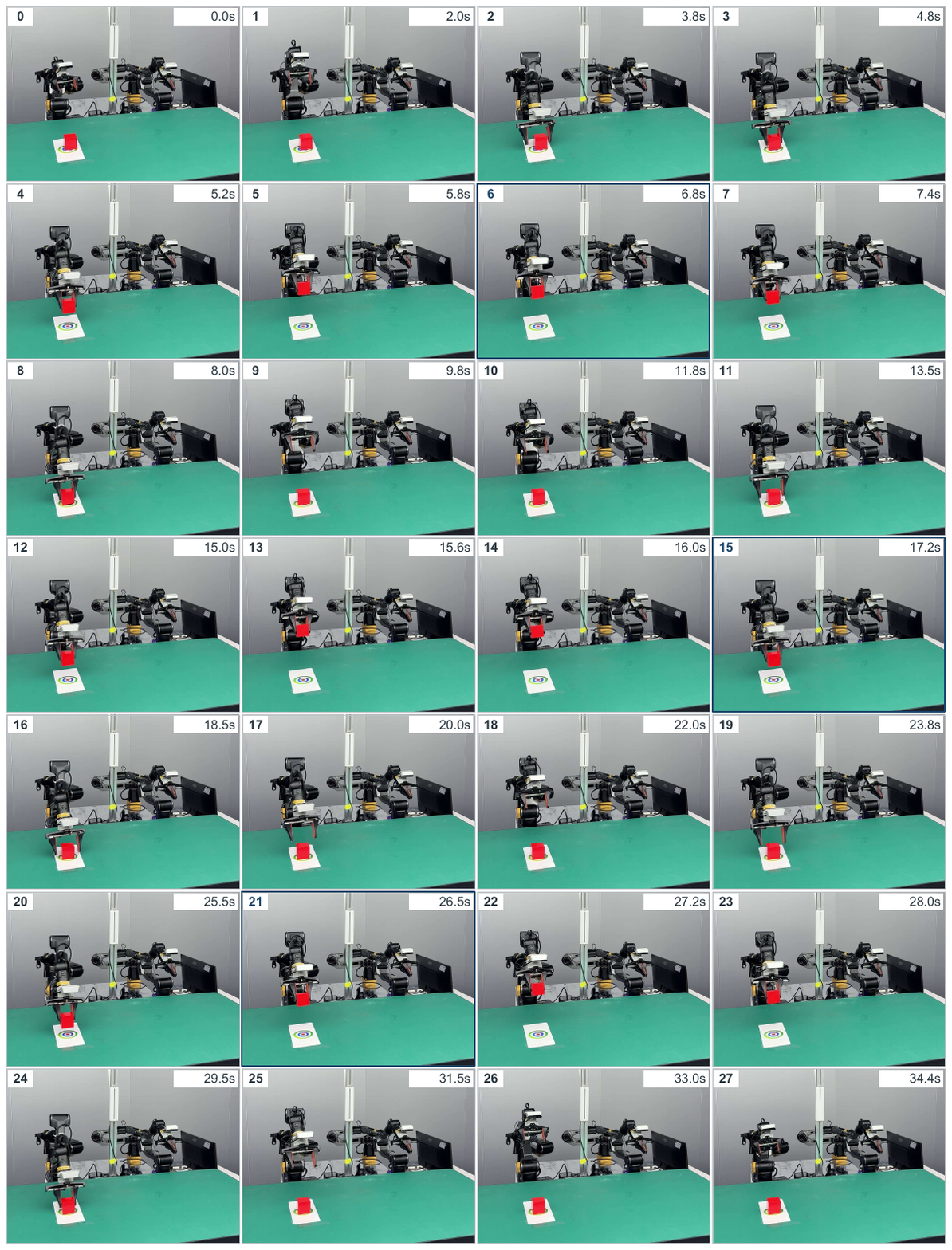}
    \caption{
        \textbf{Lift red block three times with \textsc{TaskAnchor}.}
        Across 0.0--34.4\,s, the robot performs three lift-and-return cycles with the red block. Each cycle returns the block to its marked location, producing similar tabletop observations at different cycle counts. The final frames show the block on the marker and the gripper withdrawn. Dark borders highlight representative states from the three repetitions at 6.8, 17.2, and 26.5\,s, emphasizing that similar observations occur at different execution stages.
    }
    \label{fig:app_rr_counting_taskanchor}
\end{figure}

Figure~\ref{fig:app_rr_counting_taskanchor} shows a successful \textsc{TaskAnchor} rollout. The robot completes three lift--return cycles and terminates after returning the block following the third lift. The highlighted frames emphasize similar tabletop states occurring at different repetition counts, illustrating why the current observation alone is insufficient to identify the required stage-dependent behavior.

\begin{figure}[p]
    \centering
    \includegraphics[
        width=\linewidth,
        height=0.84\textheight,
        keepaspectratio,
        trim=10bp 38bp 10bp 50bp,
        clip
    ]{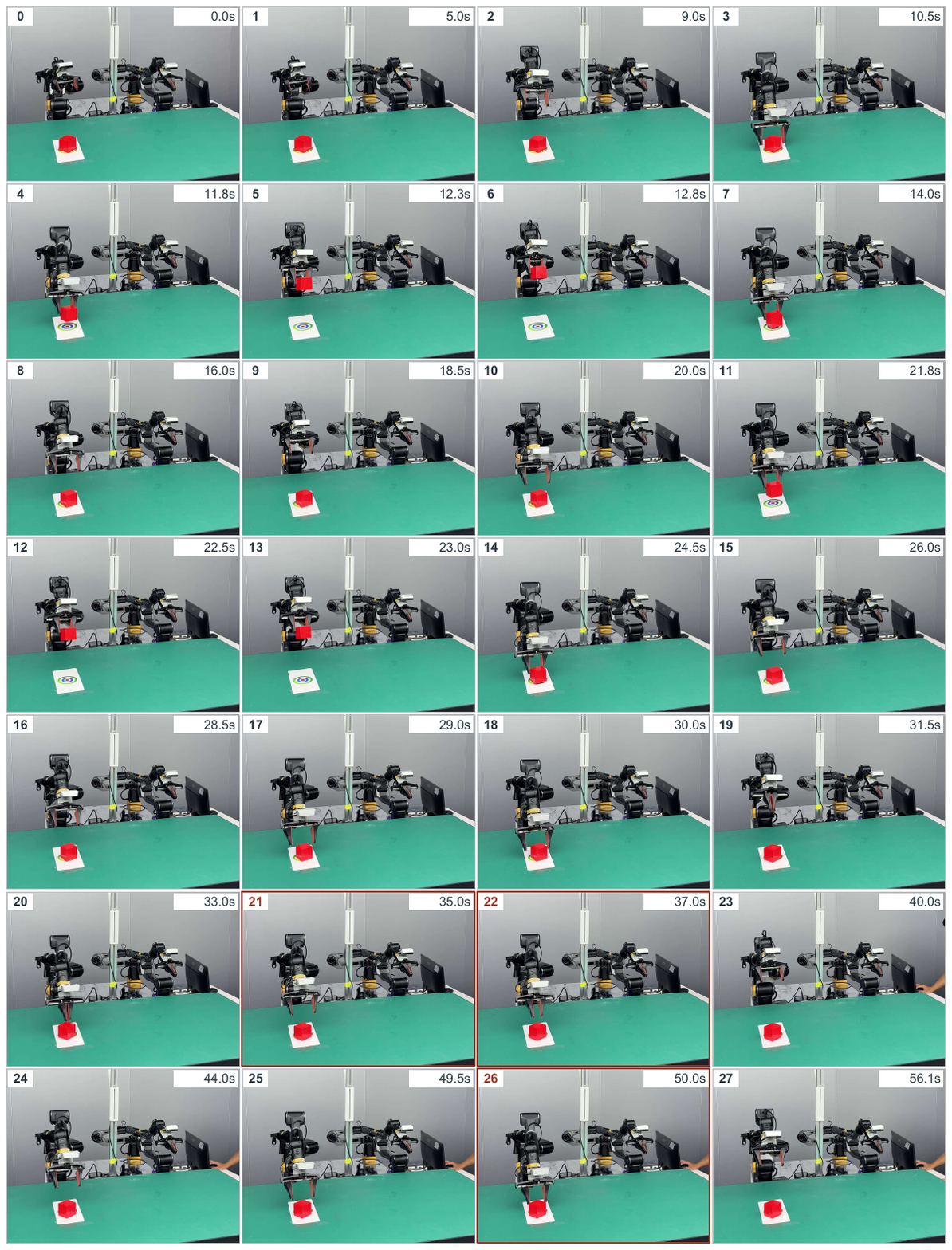}
    \caption{
        \textbf{X-VLA comparison on Lift red block three times.}
        The sequence spans 0.0--56.1\,s. Earlier frames show block lifts and returns, while later frames show repeated approaches and contacts near the marker. Later contacts do not consistently reproduce the earlier visible lift height; the images alone do not distinguish grasp-execution difficulties from an error in the retained cycle count. Red borders at 35.0, 37.0, and 50.0\,s highlight late-stage contacts and renewed approaches after the earlier lift--return cycles.
    }
    \label{fig:app_rr_counting_xvla}
\end{figure}

Figure~\ref{fig:app_rr_counting_xvla} shows a representative X-VLA comparison. Earlier frames contain successful lift and return behaviors, indicating that the baseline can execute the corresponding local manipulation primitives. Later in the rollout, however, the robot repeatedly approaches and contacts the block without reliably completing the prescribed three-cycle sequence. The selected images alone do not uniquely distinguish an error in retained repetition state from local grasp-execution difficulties, so we restrict the analysis to the visible lack of reliable stage progression and task termination.

This comparison highlights the distinction between executing an individual lift--return behavior and selecting or terminating that behavior according to the current execution stage. \textsc{TaskAnchor} conditions the policy on historical visual evidence together with the task-state coordinate, allowing otherwise similar observations to be associated with different stages of the repetitive task.

\subsubsection{Chip-Bag Checkout}
\label{app:real_robot_checkout}

The chip-bag checkout task evaluates task-state tracking across a sequence of dependent manipulation stages, including pickup, package inspection, reorientation, barcode presentation, scanning, and final placement. Successful execution requires the policy to maintain which intermediate stages have already been completed before transitioning to subsequent behaviors.

\begin{figure}[p]
    \centering
    \includegraphics[
        width=\linewidth,
        height=0.84\textheight,
        keepaspectratio,
        trim=10bp 38bp 10bp 50bp,
        clip
    ]{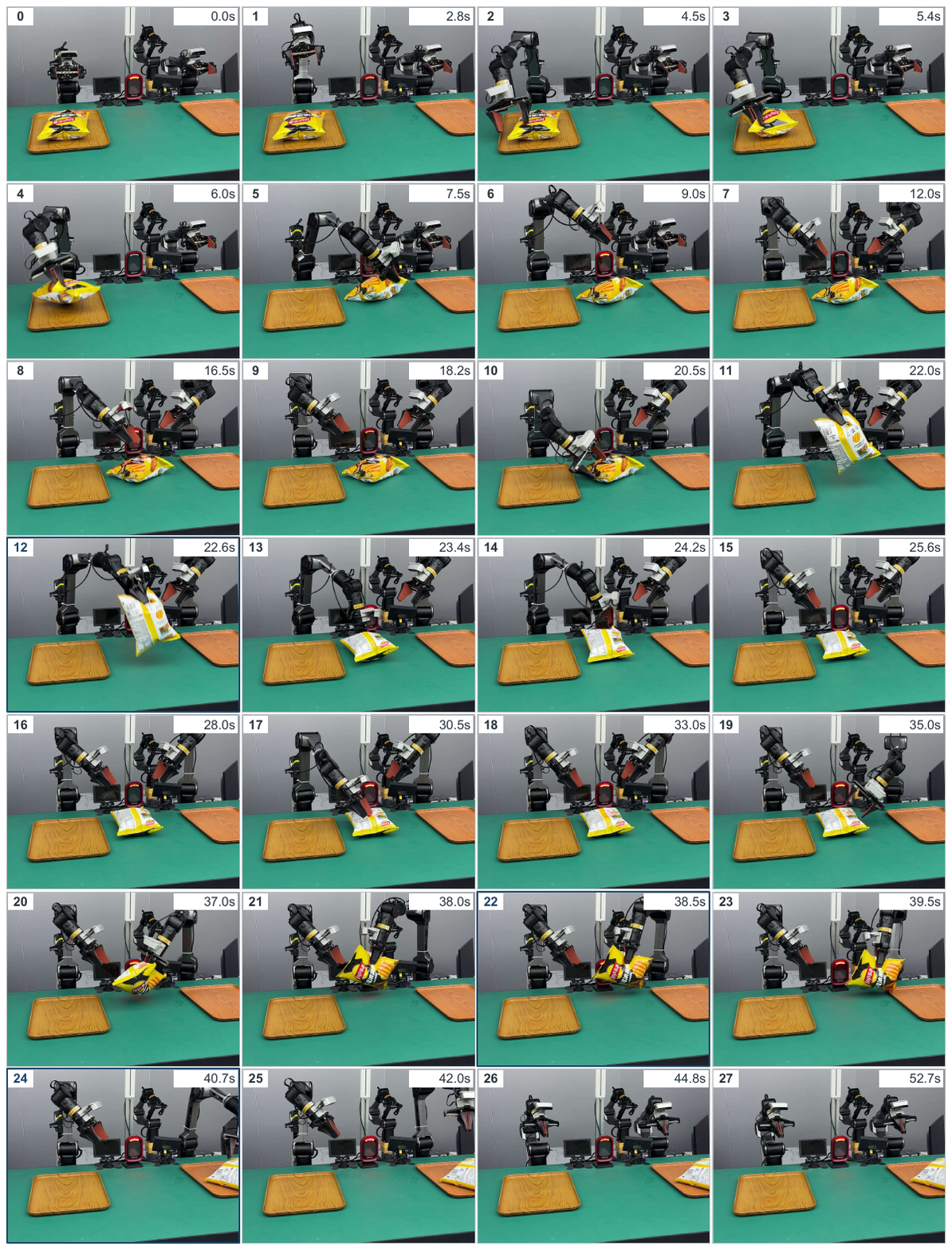}
    \caption{
        \textbf{Chip-bag checkout with \textsc{TaskAnchor}.}
        The selected frames span 0.0--52.7\,s and show pickup from the source tray, package reorientation to expose different faces, presentation near the scanner, and placement on the receiving tray. The final frames show the gripper withdrawn after the transfer. Dark borders at 22.6, 38.5, and 40.7\,s highlight package orientation and the subsequent transfer toward the receiving tray, marking progression between checkout stages.
    }
    \label{fig:app_rr_checkout_taskanchor}
\end{figure}

Figure~\ref{fig:app_rr_checkout_taskanchor} shows a successful \textsc{TaskAnchor} execution. The sequence progresses from pickup on the source tray through package reorientation and presentation near the scanner, followed by transfer to the receiving tray. Changes in the exposed package face and the subsequent transfer provide visible evidence of progression through the checkout sequence.

\begin{figure}[p]
    \centering
    \includegraphics[
        width=\linewidth,
        height=0.84\textheight,
        keepaspectratio,
        trim=10bp 38bp 10bp 50bp,
        clip
    ]{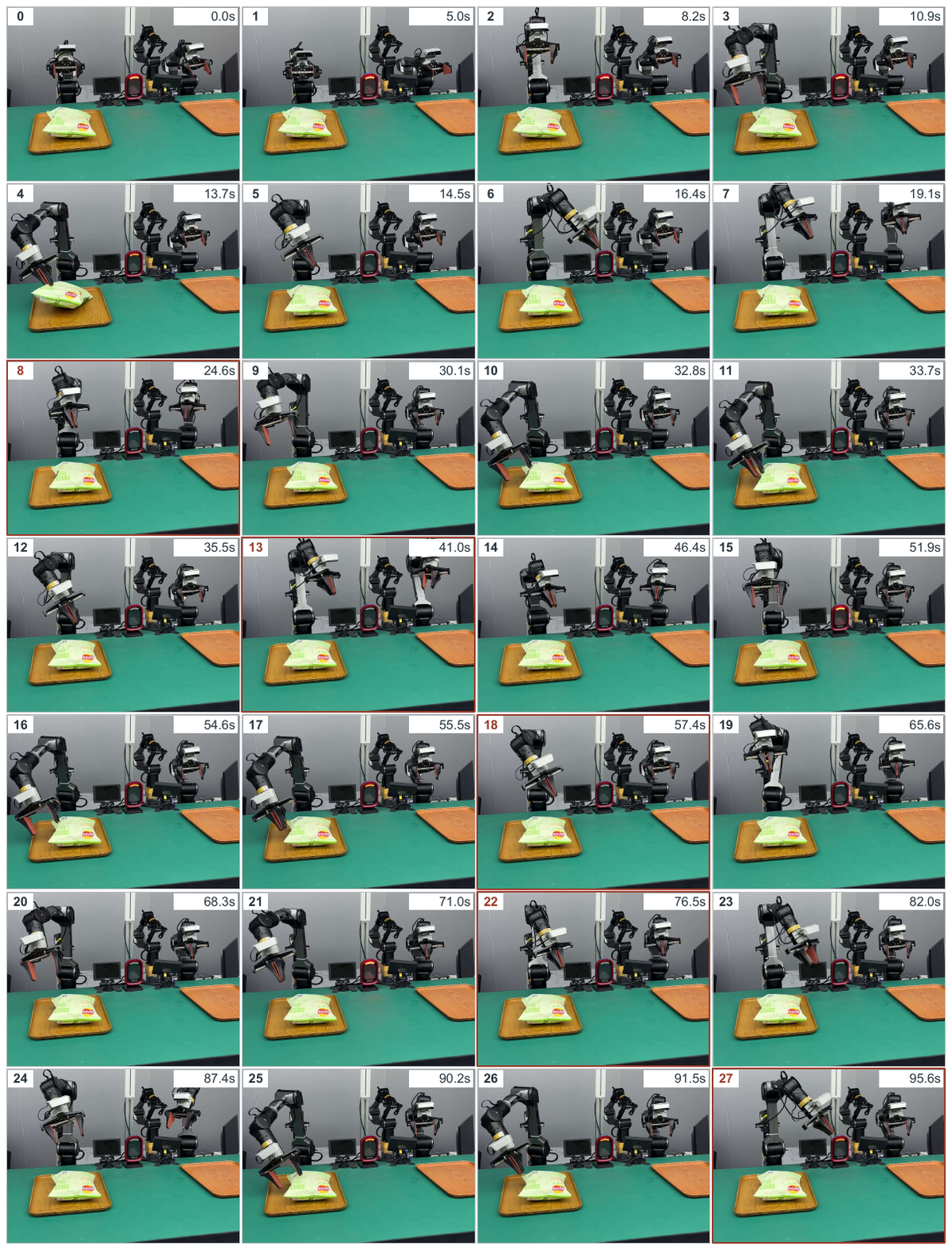}
    \caption{
        \textbf{X-VLA comparison on Chip-bag checkout.}
        Across 0.0--95.6\,s, the robot repeatedly approaches and contacts the package on the source tray. The bag remains on that tray in the final frames, and the recorded sequence does not progress to the reorientation and receiving-tray placement shown in Figure~\ref{fig:app_rr_checkout_taskanchor}. Red borders at 24.6, 41.0, 57.4, 76.5, and 95.6\,s highlight repeated pickup attempts that leave the package on the source tray.
    }
    \label{fig:app_rr_checkout_xvla}
\end{figure}

Figure~\ref{fig:app_rr_checkout_xvla} provides an X-VLA comparison. Across the recorded sequence, the baseline repeatedly approaches and contacts the package on the source tray, but the rollout does not visibly progress to the reorientation and receiving-tray placement stages observed in the \textsc{TaskAnchor} execution. The bag remains on the source tray in the final frames. Because scanner acceptance is not independently observable in the selected images, we do not infer whether individual attempts satisfy the scanning condition and restrict the comparison to visible manipulation-stage progression.

The comparison illustrates that successful long-horizon checkout requires more than executing isolated manipulation primitives: subsequent actions must be conditioned on the accumulated execution state. \textsc{TaskAnchor} provides both history-conditioned visual evidence and an explicit task-state coordinate to support these stage-dependent transitions.

\subsubsection{Human-Cued Matching}
\label{app:real_robot_human_context}

The human-cued matching task evaluates history-dependent action-branch selection under an external interaction. During each trial, a person removes and returns one reference object, after which the robot must identify and grasp the corresponding object from the bottom row. Once the object is returned and the hand withdraws, different trials can produce similar current observations even though the correct target depends on the preceding human interaction.

\begin{figure}[p]
    \centering
    \includegraphics[
        width=\linewidth,
        height=0.84\textheight,
        keepaspectratio,
        trim=10bp 38bp 10bp 50bp,
        clip
    ]{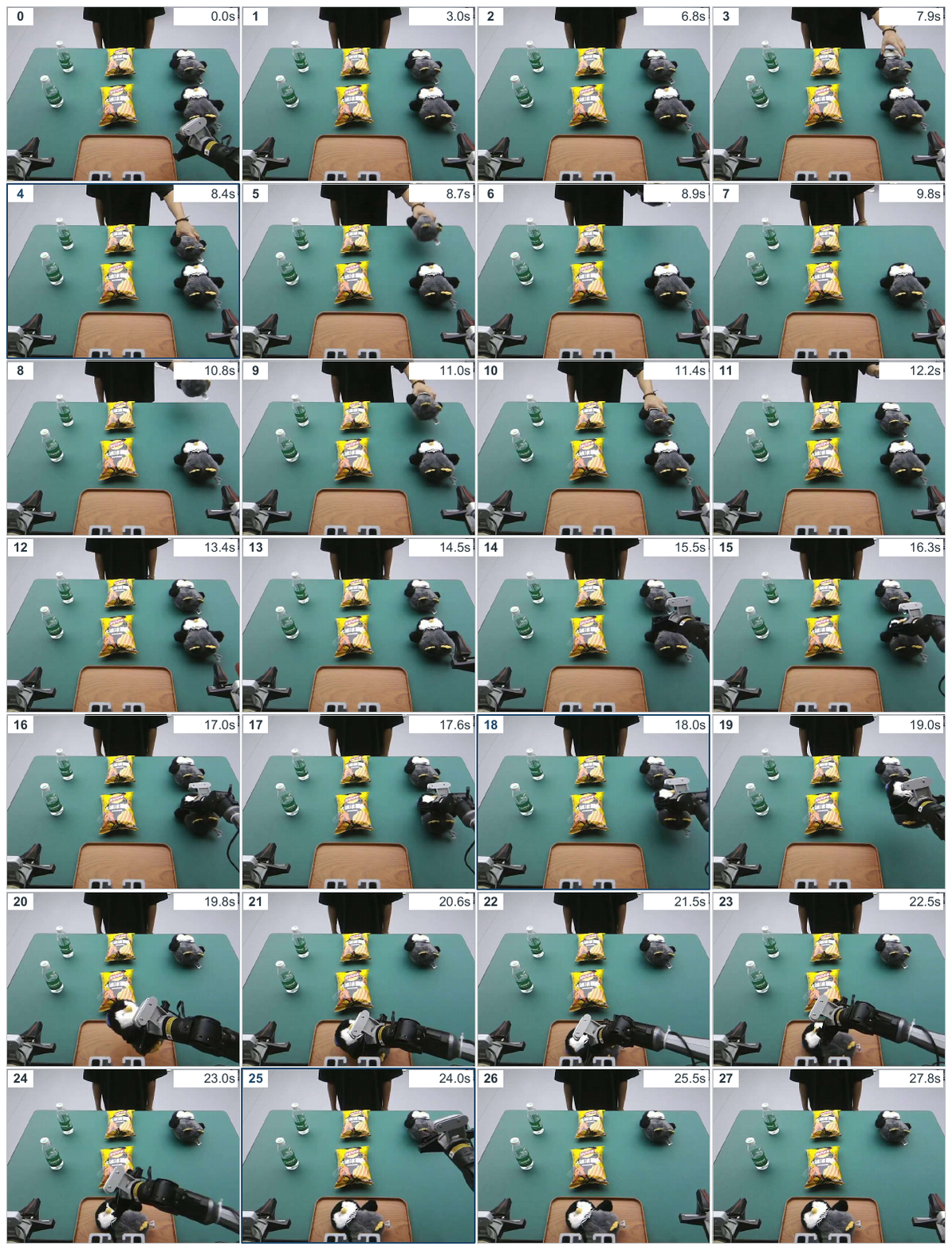}
    \caption{
        \textbf{Human-cued matching with \textsc{TaskAnchor}: plush-toy cue, egocentric view.}
        The person lifts and returns the reference plush toy. After the cue ends, the robot grasps the corresponding toy from the bottom row, transfers it to the receiving tray, and releases it. The selected frames span 0.0--27.8\,s. Dark borders highlight the human cue at 8.4\,s, the robot's matching-object grasp at 18.0\,s, and placement at 24.0\,s, linking the earlier interaction to the later object choice.
    }
    \label{fig:app_rr_human_ego_plush}
\end{figure}

\begin{figure}[p]
    \centering
    \includegraphics[
        width=\linewidth,
        height=0.84\textheight,
        keepaspectratio,
        trim=10bp 38bp 10bp 50bp,
        clip
    ]{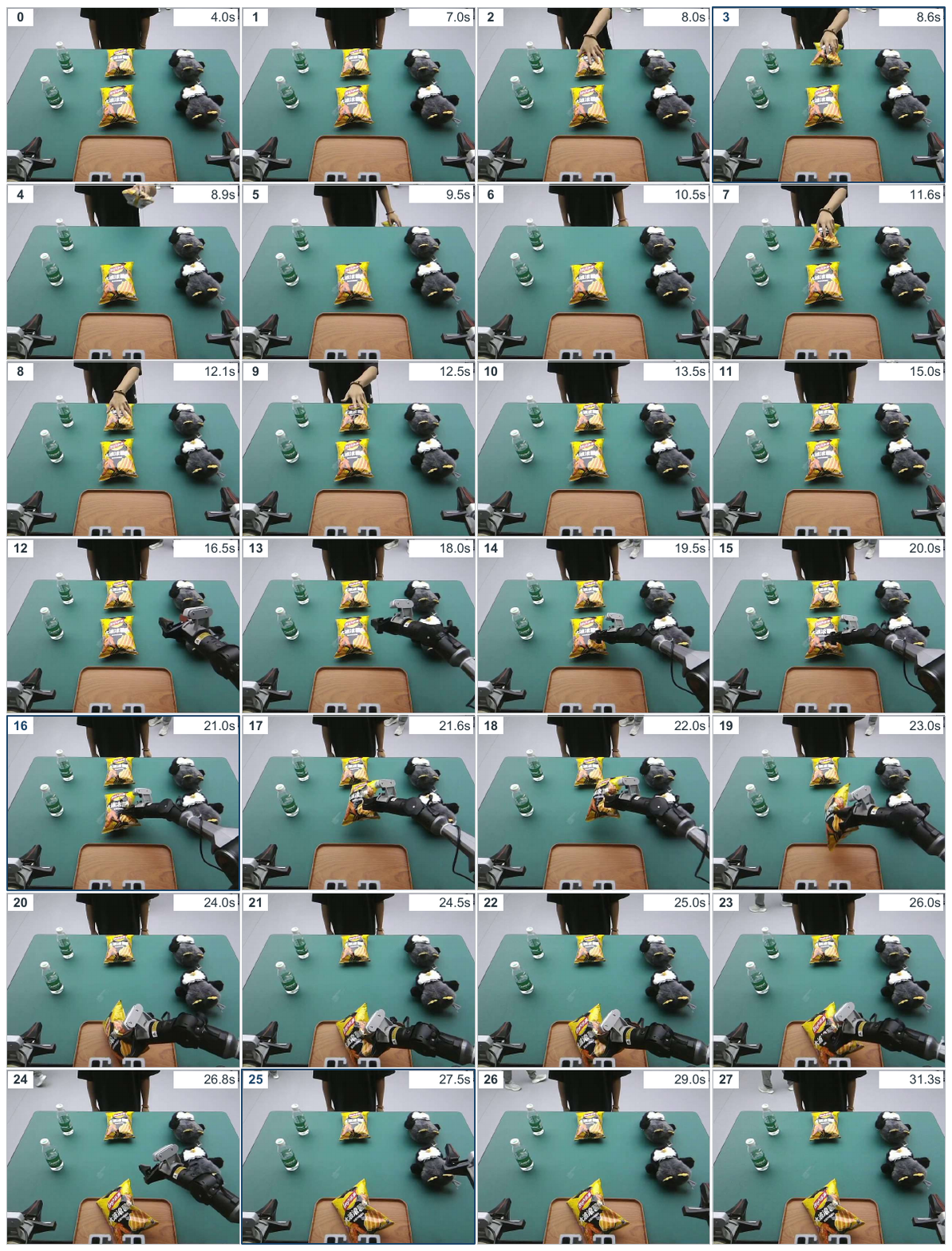}
    \caption{
        \textbf{Human-cued matching with \textsc{TaskAnchor}: chip-bag cue, egocentric view.}
        Following the lift and return of the reference chip bag, the robot selects the matching bag and deposits it on the receiving tray. Other object categories remain in the workspace. The displayed excerpt starts at 4.0\,s and ends at 31.3\,s of the source clip. Dark borders highlight the chip-bag cue at 8.6\,s, the robot's matching-bag manipulation at 21.0\,s, and placement at 27.5\,s, showing how the cue determines the selected object.
    }
    \label{fig:app_rr_human_ego_chip}
\end{figure}

\begin{figure}[p]
    \centering
    \includegraphics[
        width=\linewidth,
        height=0.84\textheight,
        keepaspectratio,
        trim=10bp 38bp 10bp 50bp,
        clip
    ]{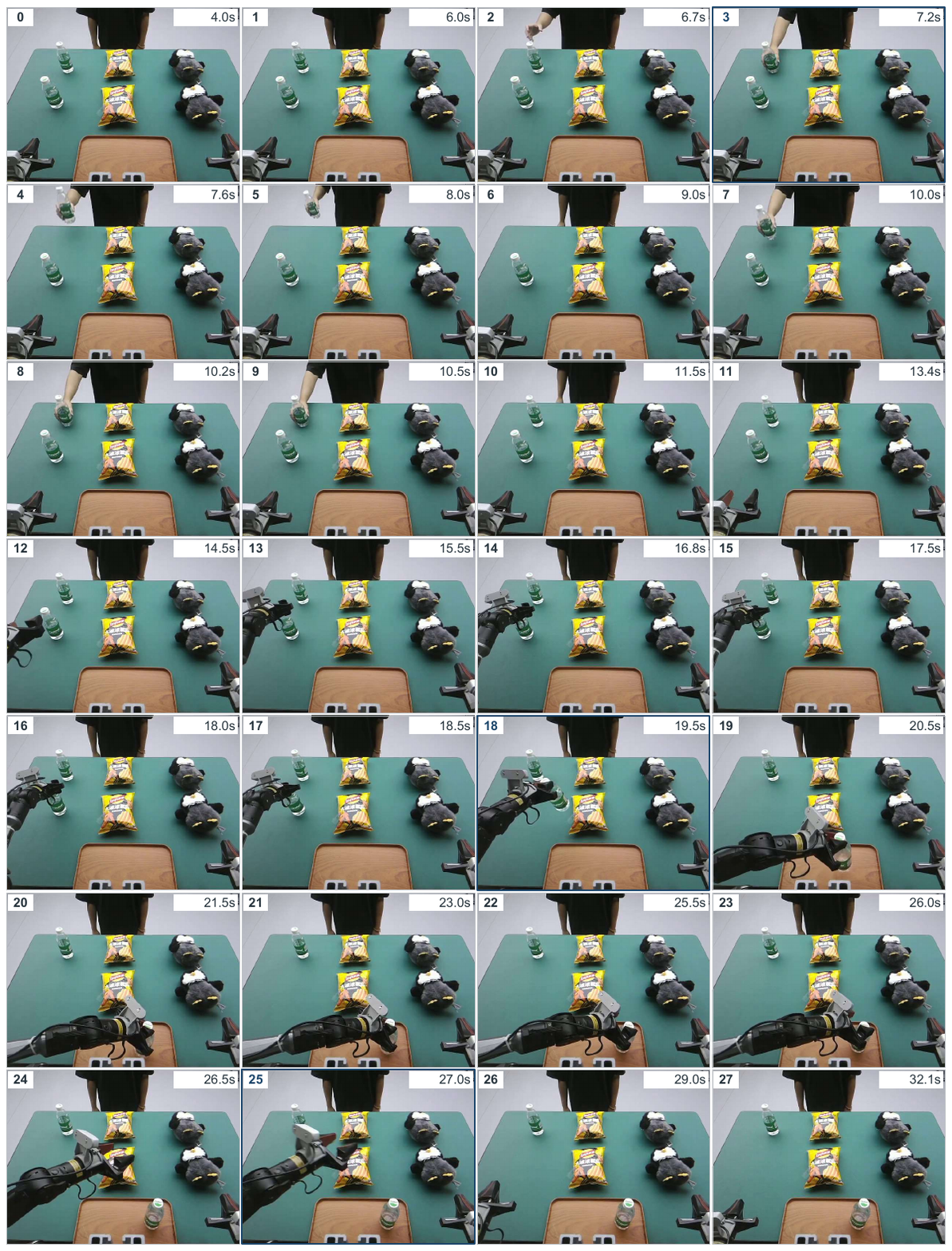}
    \caption{
        \textbf{Human-cued matching with \textsc{TaskAnchor}: bottle cue, egocentric view.}
        The person lifts and returns the reference bottle. The robot then grasps the matching bottle, transfers it toward the receiving tray, and releases it. The displayed excerpt covers 4.0--32.1\,s of the source clip and shows a different object choice from the same set of object categories. Dark borders highlight the bottle cue at 7.2\,s, the robot's matching-bottle grasp at 19.5\,s, and placement at 27.0\,s, connecting the retained cue to the bottle-selection branch.
    }
    \label{fig:app_rr_human_ego_bottle}
\end{figure}

Figures~\ref{fig:app_rr_human_ego_plush}--\ref{fig:app_rr_human_ego_bottle} show successful \textsc{TaskAnchor} executions from the egocentric view for plush-toy, chip-bag, and bottle cues, respectively. In each sequence, the person first manipulates the reference object and restores the scene. The robot then selects the corresponding object category, transfers it to the receiving tray, and releases it. Across the three examples, the subsequent target selection changes with the preceding cue despite the restored workspace.

\begin{figure}[p]
    \centering
    \includegraphics[
        width=\linewidth,
        height=0.84\textheight,
        keepaspectratio,
        trim=10bp 38bp 10bp 50bp,
        clip
    ]{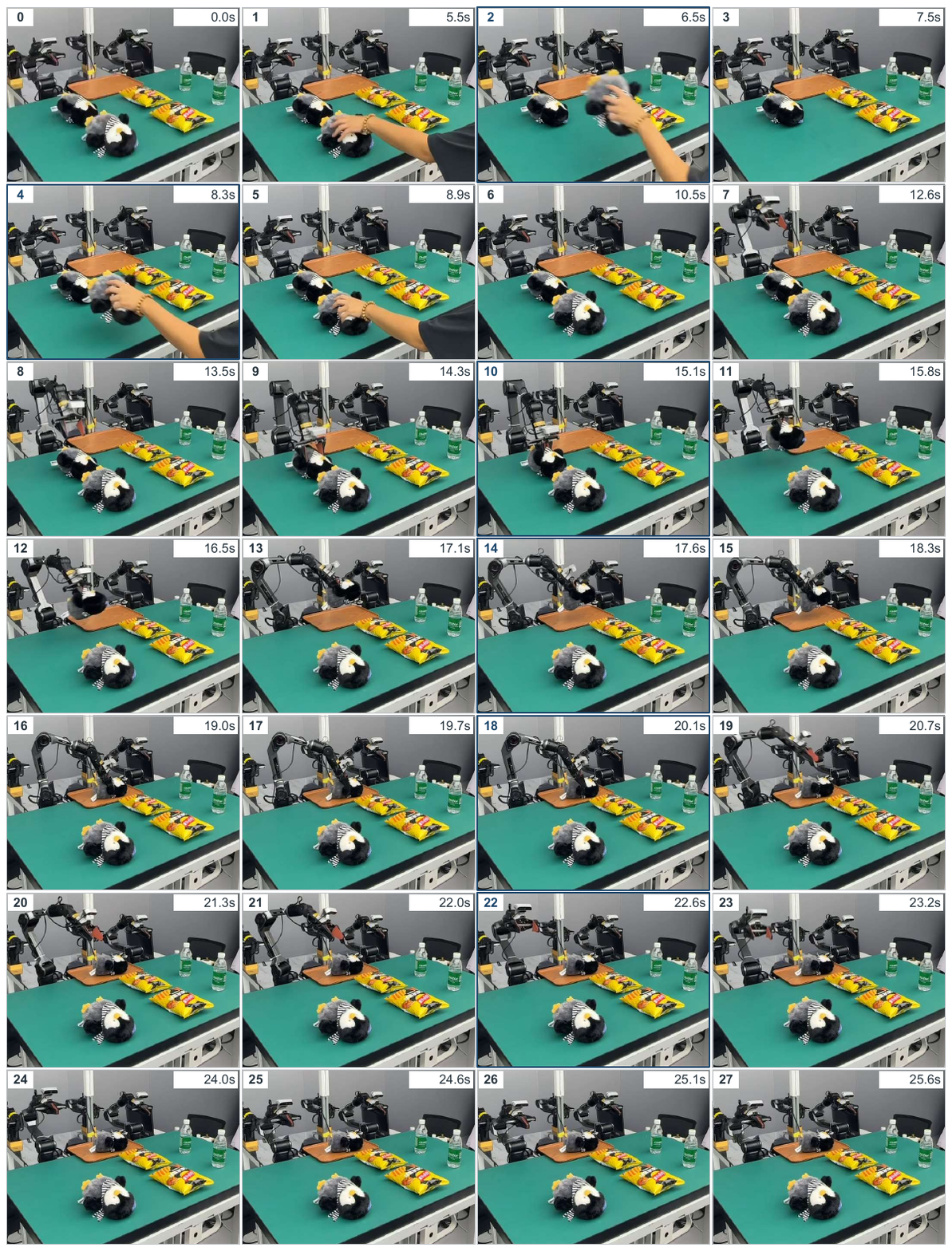}
    \caption{
        \textbf{Human-cued matching with \textsc{TaskAnchor}: plush-toy cue, external view.}
        The person lifts the reference plush toy, moves it out of view, and returns it to the table. After the hand withdraws, the robot picks the matching toy and places it on the tray. Timestamps span 0.0--25.6\,s relative to this extracted segment. Dark borders highlight cue removal and return at 6.5 and 8.3\,s, followed by representative robot manipulation states at 15.1, 17.6, 20.1, and 22.6\,s. These frames link the transient cue to the subsequent matching-object transfer and placement.
    }
    \label{fig:app_rr_human_external_plush}
\end{figure}

\begin{figure}[p]
    \centering
    \includegraphics[
        width=\linewidth,
        height=0.84\textheight,
        keepaspectratio,
        trim=10bp 38bp 10bp 50bp,
        clip
    ]{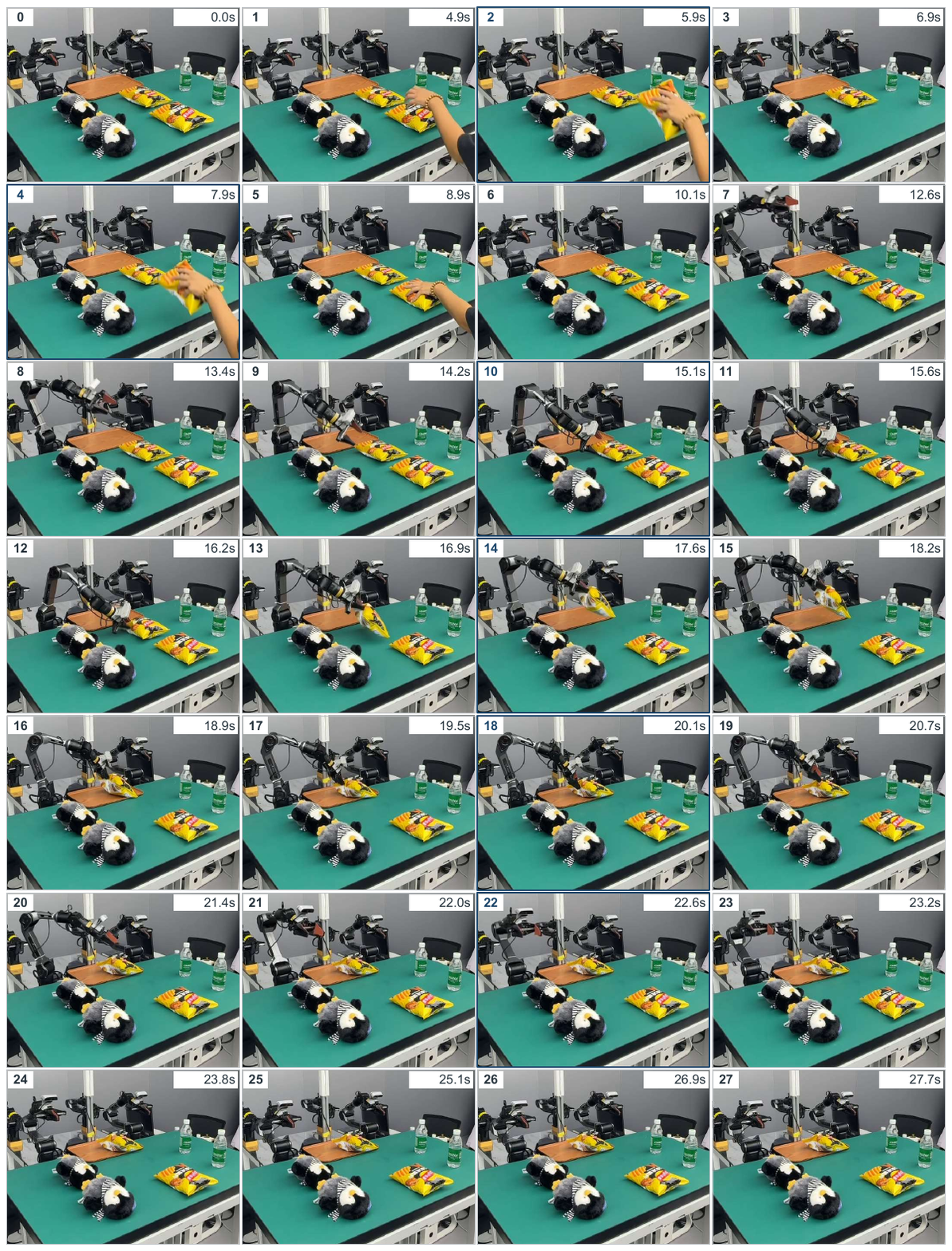}
    \caption{
        \textbf{Human-cued matching with \textsc{TaskAnchor}: chip-bag cue, external view.}
        The person lifts and returns the reference chip bag before the robot reaches for the matching bag. Subsequent frames show grasp, transfer, release onto the tray, and gripper withdrawal. Timestamps span 0.0--27.7\,s relative to this extracted segment. Dark borders highlight cue removal and return at 5.9 and 7.9\,s, together with manipulation states at 15.1, 17.6, 20.1, and 22.6\,s, tracing the transition from the human cue to the matching bag's transfer and placement.
    }
    \label{fig:app_rr_human_external_chip}
\end{figure}

\begin{figure}[p]
    \centering
    \includegraphics[
        width=\linewidth,
        height=0.84\textheight,
        keepaspectratio,
        trim=10bp 38bp 10bp 50bp,
        clip
    ]{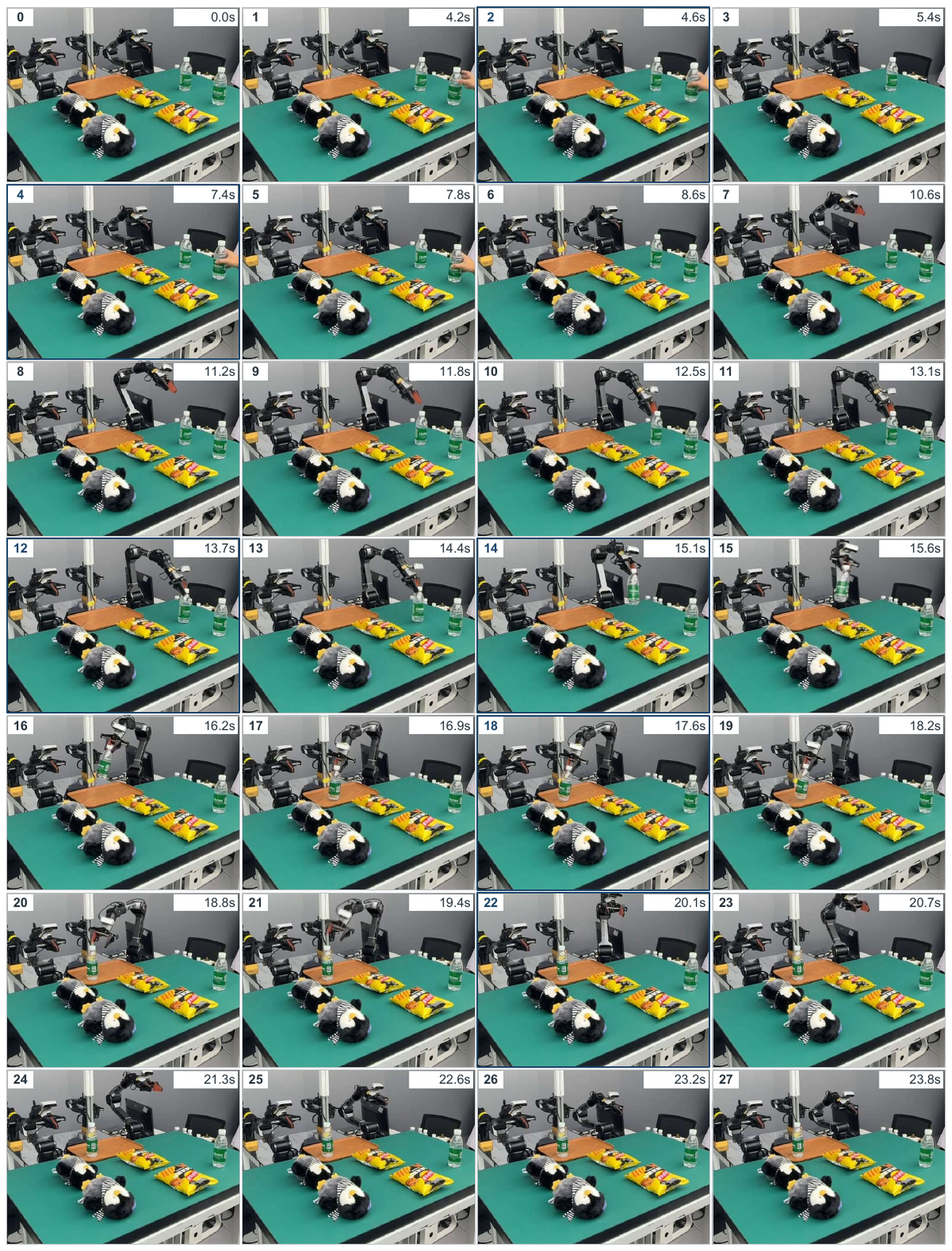}
    \caption{
        \textbf{Human-cued matching with \textsc{TaskAnchor}: bottle cue, external view.}
        The person removes and returns the reference bottle. After the cue ends, the robot selects the matching bottle, places it upright on the tray, and withdraws. Timestamps span 0.0--23.8\,s relative to this extracted segment. Dark borders highlight cue removal and return at 4.6 and 7.4\,s and subsequent robot manipulation at 13.7, 15.1, 17.6, and 20.1\,s, showing progression from the bottle cue through grasp, transfer, and placement.
    }
    \label{fig:app_rr_human_external_bottle}
\end{figure}

Figures~\ref{fig:app_rr_human_external_plush}--\ref{fig:app_rr_human_external_bottle} show the same task categories from an external view. These sequences are separate recordings rather than synchronized views of the egocentric rollouts. They retain the same task structure: cue removal and return are followed by robot approach, matching-object grasp, transfer, and placement.

\begin{figure}[p]
    \centering
    \includegraphics[
        width=\linewidth,
        height=0.84\textheight,
        keepaspectratio,
        trim=10bp 38bp 10bp 50bp,
        clip
    ]{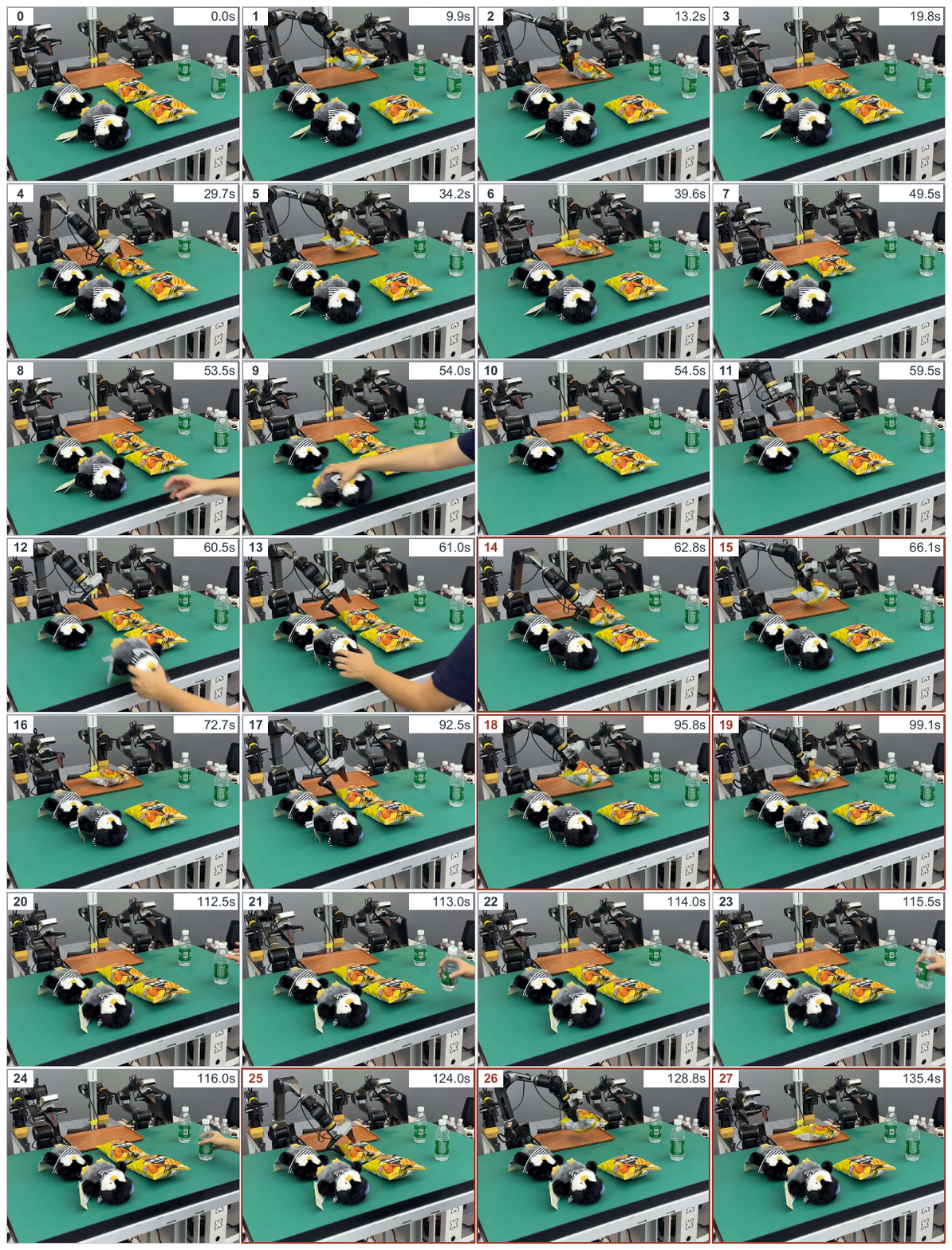}
    \caption{
        \textbf{X-VLA comparison on Human-cued matching.}
        The sequence spans 0.0--135.4\,s. Following earlier chip-bag transfers, the person removes and returns the plush toy at approximately 54--61\,s and later the bottle at approximately 113--116\,s. The robot subsequently continues to transfer the chip bag, showing an object-selection mismatch with the changed human cue. Red borders highlight continued chip-bag manipulation at 62.8, 66.1, 95.8, and 99.1\,s after the plush-toy cue, and at 124.0, 128.8, and 135.4\,s after the bottle cue. The selected frames expose repeated object choices inconsistent with the most recent human interaction.
    }
    \label{fig:app_rr_human_xvla}
\end{figure}

Figure~\ref{fig:app_rr_human_xvla} shows a representative X-VLA comparison. After earlier chip-bag interactions, the human subsequently provides plush-toy and bottle cues, yet the robot continues to manipulate the chip bag. The selected frames therefore expose a visible cue--action mismatch: the object selected by the robot is inconsistent with the most recent human interaction. This observation does not by itself identify the baseline's internal failure mechanism, but it shows that its subsequent object choice is not reliably adapted to the changed external cue.

This task directly exposes history dependence that cannot be resolved from the restored scene alone. Once the transient human interaction has disappeared from the current observation, the correct target still depends on that earlier event. \textsc{TaskAnchor} retains the interaction history as temporal context, allowing the subsequent action branch to be conditioned on the external cue rather than only on the final visual arrangement.

\paragraph{Summary of Real-Robot Failure Patterns.}
Together, these comparisons illustrate three forms of history-dependent action ambiguity in physical manipulation. Repeated lifting requires distinguishing visually similar states at different repetition stages; chip-bag checkout requires maintaining progression through dependent intermediate stages; and human-cued matching requires retaining a transient external interaction that determines a later action branch. Across these scenarios, the X-VLA baseline can exhibit useful local manipulation behaviors but does not reliably organize them according to the hidden execution context required by the full task. \textsc{TaskAnchor} addresses this ambiguity by conditioning the pretrained policy on both historical visual evidence and an explicit task-state coordinate.

\clearpage

\clearpage

\clearpage
\section{Additional Ablations}
\label{app:additional_ablations}

\subsection{Task-State Coordinate Analysis}
\label{app:coordinate_ablation}

We first examine the sensitivity of \textsc{TaskAnchor} to the coordinate-supervision weight $\lambda_{\mathrm{coord}}$ in Eq.~(8). We evaluate $\textbf{\textsc{TaskAnchor}-XVLA}^{*}$ on the five RMBench tasks with $\lambda_{\mathrm{coord}}\in\{2,1,0.5\}$ while keeping all other training and evaluation settings fixed. Each task is evaluated over 10 episodes, yielding 50 rollouts for each setting.

\begin{table}[!htbp]
    \centering
    \caption{
        \textbf{Sensitivity to the task-state coordinate loss weight on RMBench.}
        All variants use $\textbf{\textsc{TaskAnchor}-XVLA}^{*}$ and differ only in $\lambda_{\mathrm{coord}}$.
        Each setting is evaluated over 10 episodes per task, for a total of 50 rollouts across the five RMBench tasks.
    }
    \label{tab:lambda_coord_ablation}
    \begin{tabular}{c c c}
        \toprule
        $\lambda_{\mathrm{coord}}$ & Successful Episodes & Overall SR \\
        \midrule
        $2$ & $32/50$ & $0.64$ \\
        $\mathbf{1}$ & $\mathbf{34/50}$ & $\mathbf{0.68}$ \\
        $0.5$ & $31/50$ & $0.62$ \\
        \bottomrule
    \end{tabular}
\end{table}

As shown in Table~\ref{tab:lambda_coord_ablation}, \textsc{TaskAnchor} remains effective across the tested coordinate-loss weights, with overall success rates ranging from 0.62 to 0.68. The default setting $\lambda_{\mathrm{coord}}=1$ achieves the highest success rate in this evaluation, with 34 successful episodes out of 50, compared with 32/50 for $\lambda_{\mathrm{coord}}=2$ and 31/50 for $\lambda_{\mathrm{coord}}=0.5$. We therefore use $\lambda_{\mathrm{coord}}=1$ throughout the main experiments. The moderate variation across these settings suggests that the method does not depend on a narrowly tuned coordinate-loss weight, while maintaining a balanced contribution of coordinate supervision to the joint training objective.

\begin{figure}[!htbp]
    \centering
    \includegraphics[width=\linewidth]{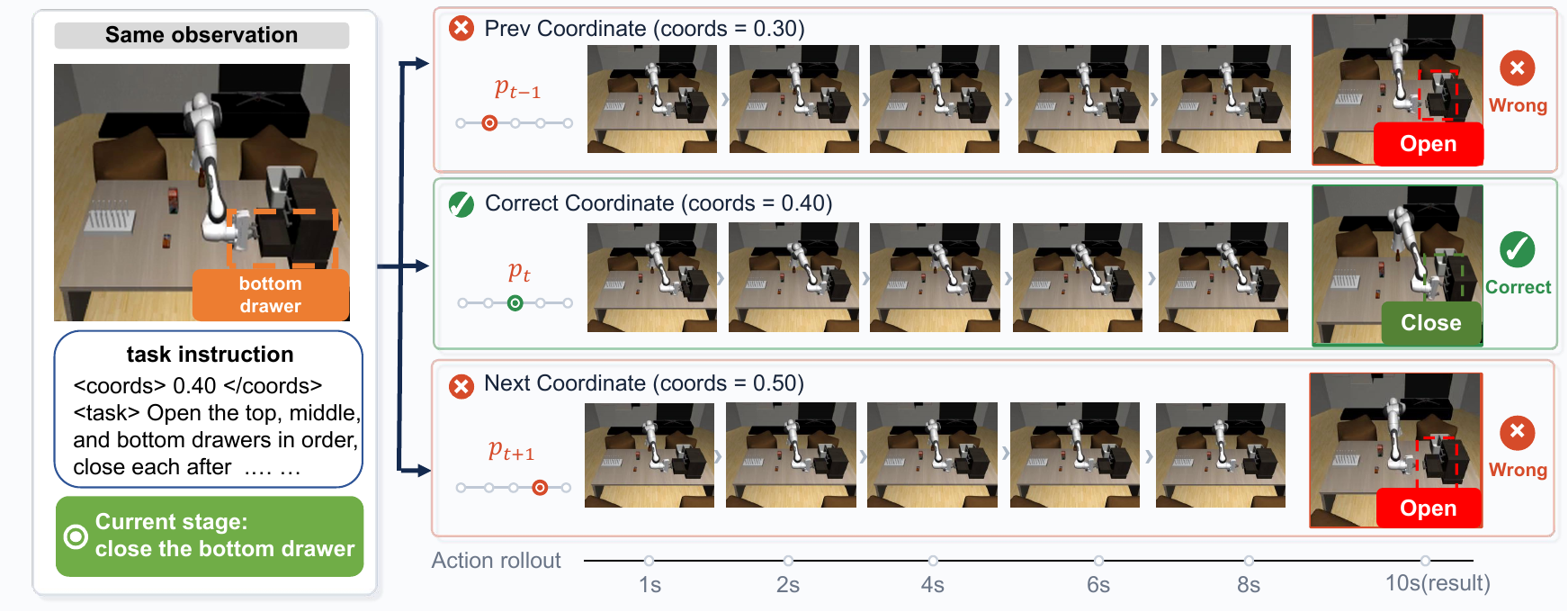}
    \caption{
        \textbf{Controlled intervention on the task-state coordinate.}
        Given the same visual observation and language instruction, the correct coordinate ($p_t=0.40$) makes \taskanchorpi{} close the bottom drawer, whereas the preceding ($p_t=0.30$) or subsequent ($p_t=0.50$) coordinate induces an incorrect drawer-opening behavior. Only the task-state coordinate is changed across the three rollouts.
    }
    \label{fig:task_state_intervention}
\end{figure}

We further examine whether the task-state coordinate directly affects action selection through a controlled intervention experiment. As shown in Figure~\ref{fig:task_state_intervention}, we keep the current visual observation and language instruction fixed and vary only the injected task-state coordinate. The current execution stage requires the robot to close the bottom drawer, and the correct coordinate is $p_t=0.40$.

With the correct coordinate, \taskanchorpi{} generates the appropriate closing behavior and completes the current stage. In contrast, replacing the coordinate with the preceding value ($p_t=0.30$) or the subsequent value ($p_t=0.50$) changes the generated rollout toward an incorrect drawer-opening behavior, despite identical visual and language inputs. This controlled intervention isolates the effect of the task-state coordinate from changes in observation or instruction.

The intervention shows that the task-state coordinate is actively used as an execution-state condition rather than serving only as an auxiliary prediction target. Nearby but incorrect coordinates induce behaviors associated with different execution stages, whereas the correct coordinate recovers the behavior required by the current stage. Together with the loss-weight sensitivity analysis, these results show that the coordinate branch provides a stable supervised task-state signal during training and directly conditions stage-dependent action selection at inference time.

\section{Limitations and Future Directions}
\label{app:limitations}

Although \textsc{TaskAnchor} substantially improves long-horizon manipulation by
providing task-state-aware conditions to pretrained VLAs, several
limitations remain and motivate future research.

First, the current task-state coordinate relies on milestone-based
supervision during post-training. This design provides an effective and
lightweight way to ground execution stages without introducing additional
planning modules, but it requires task-specific semantic stage
annotations. Future work could investigate automatically discovering
task-state representations from large-scale robot interaction data,
allowing execution states to emerge from experience rather than manually
defined milestones.

Second, \textsc{TaskAnchor} demonstrates that pretrained VLAs can acquire
history-aware and context-dependent behaviors through lightweight
post-training. However, the extent to which this learned context
adaptation generalizes beyond the demonstrated task distribution remains
an open question. In particular, whether task-state representations
learned from observed manipulation trajectories can transfer to completely
unseen tasks, novel object configurations, or previously unobserved
execution structures has not been extensively evaluated in this work.
Future research will explore more general task-state representations that
enable broader in-context adaptation across diverse manipulation
scenarios.

Third, \textsc{TaskAnchor} currently models task state within an individual
execution trajectory. Extending this formulation toward longer-term
robotic experience \citep{yue2025real}, where task states evolve across multiple tasks,
sessions, and environments, remains an important direction. Future work
could investigate persistent task-state memory and continual adaptation
mechanisms that allow robots to accumulate and reuse execution knowledge
over extended periods of interaction.

\section{Additional Visualization of Ground-Truth Task-State Coordinates}
\label{app:coordinate_gt}

To further illustrate how the task-state coordinate is annotated, we provide
two additional trajectory visualizations from RMBench\citep{chen2026rmbench} and RoboMemArena~\citep{lei2026robomemarena}.

\begin{figure}[p]
    \centering
    \includegraphics[width=\linewidth]{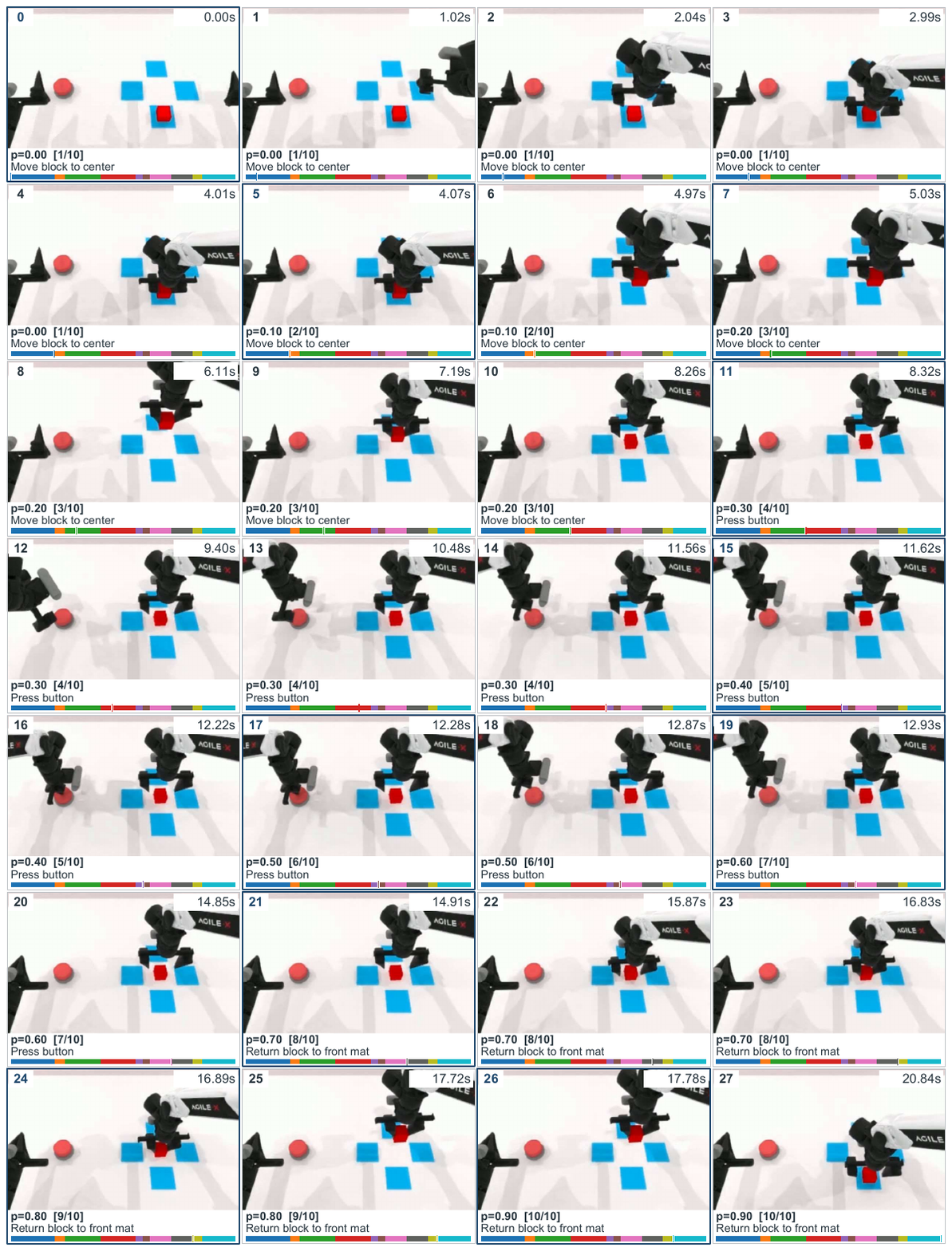}
    \caption{
        \textbf{Ground-truth task-state coordinate annotation on RMBench.}
        Visualization of a full \textit{Put Back Block} trajectory with
        milestone-level task-state coordinate labels.
        This example contains ten semantic milestones with
        $p_t^\star \in \{0.0, 0.1, \ldots, 0.9\}$.
        Although several consecutive milestones share the same high-level
        behavior label, such as \emph{Move block to center},
        \emph{Press button}, or \emph{Return block to front mat}, they are
        assigned distinct coordinates according to their execution stage.
        This illustrates that the task-state coordinate captures semantic
        execution state rather than subtask name alone.
    }
    \label{fig:coordinate_gt_rmbench}
\end{figure}

\begin{figure}[p]
    \centering
    \includegraphics[width=\linewidth]{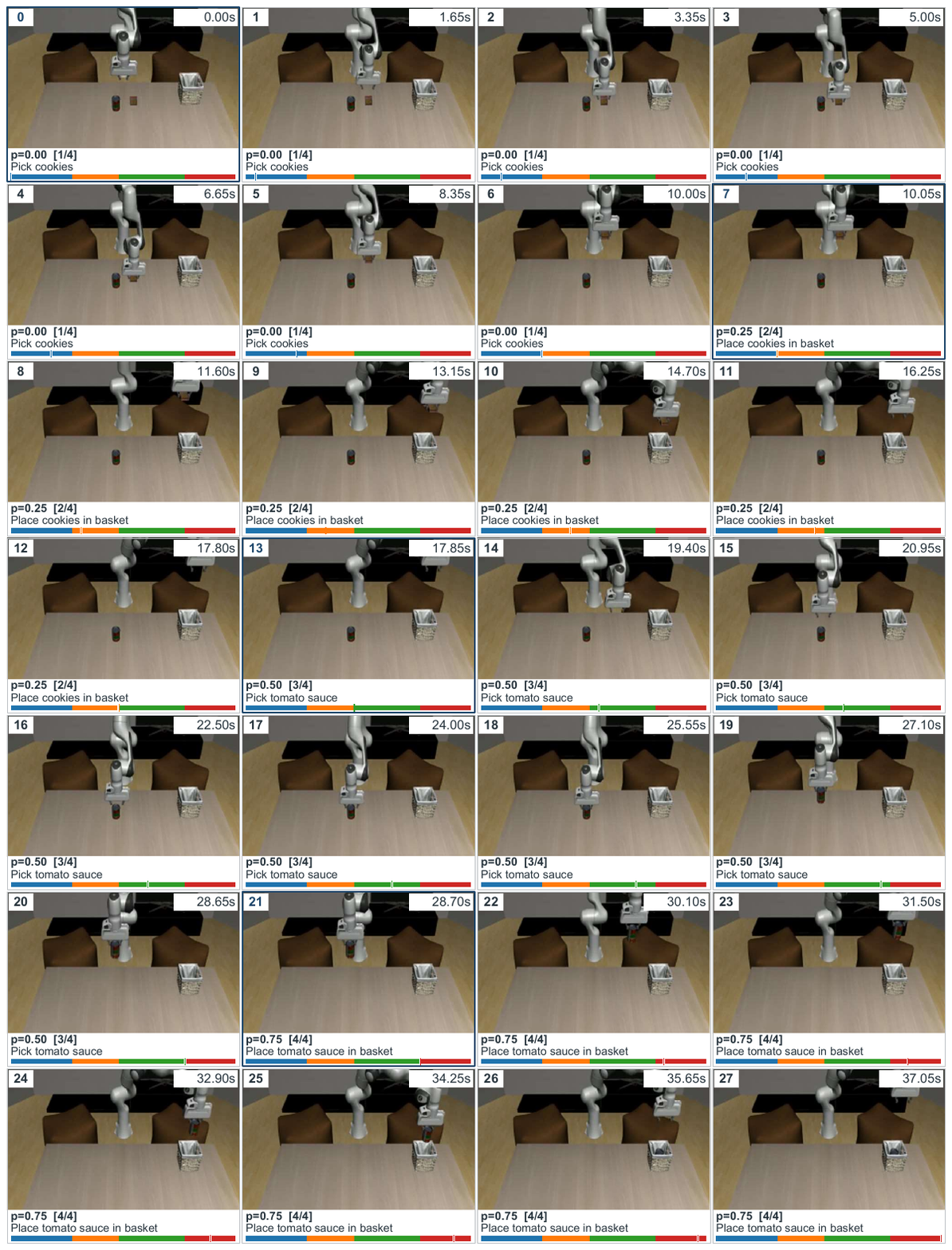}
    \caption{
        \textbf{Ground-truth task-state coordinate annotation on RoboMemArena.}
        Visualization of a full multi-sequence trajectory with milestone-level
        task-state coordinate labels.
        This example contains four semantic milestones with
        $p_t^\star \in \{0.00, 0.25, 0.50, 0.75\}$, corresponding to
        \emph{Pick cookies}, \emph{Place cookies in basket},
        \emph{Pick tomato sauce}, and
        \emph{Place tomato sauce in basket}, respectively.
        Compared with the RMBench example, this trajectory has a different
        milestone structure and temporal length, illustrating that the
        task-state coordinate is normalized by task-specific milestone
        annotations rather than raw trajectory duration.
    }
    \label{fig:coordinate_gt_robomem}
\end{figure}

Figure~\ref{fig:coordinate_gt_rmbench} shows a full annotated trajectory from
RMBench \textit{Put Back Block}, while
Figure~\ref{fig:coordinate_gt_robomem} shows a full annotated trajectory from a
RoboMemArena multi-sequence task.

\end{document}